\documentclass{article}

\usepackage{arxiv}

\usepackage[utf8]{inputenc} 
\usepackage[T1]{fontenc}    
\usepackage{hyperref}       
\usepackage{url}            
\usepackage{booktabs}       
\usepackage{amsfonts}       
\usepackage{nicefrac}       
\usepackage{microtype}      
\usepackage{lipsum}
\usepackage{subfigure}
\usepackage{graphicx}
\usepackage[ruled,vlined,noresetcount]{algorithm2e}
\usepackage{amsmath}

\title{Long-Bone Fracture Detection using Artificial Neural Networks based on Line Features of X-ray Images}

\author{
  Alice Yi Yang \\ 
  School of Electrical and Information Engineering \\
  University of the Witwatersrand \\
  Johannesburg, South Africa, 2000 \\
  \texttt{yangalice8@gmail.com} \\
   \And
 Ling Cheng \\
  School of Electrical and Information Engineering \\
  University of the Witwatersrand \\
  Johannesburg, South Africa, 2000 \\
  \texttt{ling.cheng@wits.ac.za} \\
}

\begin{document}
\maketitle

\begin{abstract}
Two line-based fracture detection scheme are developed and discussed, namely Standard line-based fracture detection and Adaptive Differential Parameter Optimized (ADPO) line-based fracture detection. The purpose for the two line-based fracture detection schemes is to detect fractured lines from X-ray images using extracted features based on recognised patterns to differentiate fractured lines from non-fractured lines. The difference between the two schemes is the detection of detailed lines. The ADPO scheme optimizes the parameters of the Probabilistic Hough Transform, such that granule lines within the fractured regions are detected, whereas the Standard scheme is unable to detect them. The lines are detected using the Probabilistic Hough Function, in which the detected lines are a representation of the image edge objects. The lines are given in the form of points, $(x,y)$, which includes the starting and ending point. Based on the given line points, 13 features are extracted from each line, as a summary of line information. These features are used for fracture and non-fracture classification of the detected lines. The classification is carried out by the Artificial Neural Network (ANN). There are two evaluations that are employed to evaluate both the entirety of the system and the ANN. The Standard Scheme is capable of achieving an average accuracy of 74.25\%, whilst the ADPO scheme achieved an average accuracy of 74.4\%. The ADPO scheme is opted for over the Standard scheme, however it can be further improved with detected contours and its extracted features.
\end{abstract}


\section{Introduction}
The application of Computer Aided Diagnosis (CAD) in the medical field has been introduced since the early 1970s \cite{potter1988method, nahar_liver_2018}, in which the first CAD system utilised a decision tree analysis. Since the early 1970s, CAD systems have developed further and some even employ the use of Artificial Neural Networks (ANN). Syiam, M Et al. \cite{syiam_adagen:_2004} proposed an adaptive interface agent (AdAgen) for X-ray fracture detection. The interface AdAgen uses neural network to collaborate with trained agents. The neural network is used to build the software interface agent for the detection of fractures in long bones. A semi-intelligent system is provided by the software agent. The results obtained from the simulations indicates that the incorporated agents assists with the performance of the automated fracture detection in leg radiography. The general approach to classifying the presence of bone fracture involves mapping the data to one of several predefined classes. However, there are challenges presented in the classification techniques, which are due to information overload, size and dimension of the data \cite{Mahendran2011}. A classification technique is defined as a systematic approach of processing input data by constructing classification models. Examples of classification techniques includes Decision Tree Classifiers, Rule-Based Classifiers, Neural Networks, Support Vector Machines and Na\"{i}ve Bayes Classifiers. The authors of \cite{mahendran_Enhanced}, proposes a four-step system that makes use of fusion-classification techniques to automate the detection of bone fracture specifically for leg bones (tibia). The four-steps includes preprocessing, segmentation, feature extraction and bone detection. The three classifiers during the fusion classification are Back-Propagation Neural Network (BPNN), Support Vector Machine (SVM) and Na\"{i}ve Bayes Classifiers (NB). Through experimentation, the authors stated that the proposed four-step system showed significant improvement in terms of detection rate and speed of classification. An alternative classifier is the Convolutional Neural Network (CNN) which is a supervised classifier. Do\u{g}antekin A. Et al. \cite{dogantekin_novel_2019} proposed a hybrid system with a CNN and Wavelet Transform-Singular Value Decomposition (DWT-SVD) for the classification of malignant and benign masses from liver CT images. The authors intension for the hybrid system is to reduce the execution time of the CNN architecture. The features for the classification are extracted using Perceptual hash functions. The hybrid system was evaluated using 200 images, in which 100 images are of benign tumours and the other 100 are of malignant tumours. The system achieved an accuracy of 97.3\%. Features are crucial for the classification of various categories. The authors of \cite{hou_deep_2019} developed a model based on Deep Convolutional Neural Network (DCNN) for feature extractions. The features are extracted from X-ray images for the classification of weld flaw types. The model uses a technique called Sparse Autoencoder (SAE) as well as the mapping of the CNN layers for feature extraction. This extracted features are compared to the features from the traditional Grey Level Co-Occurrence Matrix (GLCM) technique. The results obtained by the authors indicate that the extracted features based on the DCNN is better than traditional features, as it obtained a 97.2\% accuracy whereas the accuracy obtained by the traditional methods obtained an 82.2\% accuracy. The processing of images are essential as it ensures that all images are consistent to obtain the best results when passed through the system. In \cite{xing_gradation_2019} the proposed system employs fuzzy network, multilevel threshold and morphological methods for the removal of noise and enhance the contrast of the CT images. The fuzzy network proved to balance the noise reduction and enhancement in the results, whilst the multilevel Otsu's threshold filled in the missing gaps found within the images.

This paper describes a novel Adaptive Differential Parameter Optimization (ADPO) line-based fracture detection scheme. The purpose of the line-based fracture detection scheme is to offer a second opinion to medical physicians by classifying fractured and non-fractured lines obtained from X-ray images. The proposed novel technique is intended classify fractured lines from non-fractured lines, by training the Artificial Neural Network (ANN) with lines rather than images. Consequently, this reduces the number of images required for training to achieve an accuracy above 70\%. In order to understand the ADPO approach for line-based fracture detection, the Standard line-based fracture detection is detailed in Section \ref{Section: Standard Line-Based Fracture Detection}. The Standard line-based fracture detection procedure extracts lines from the canny generated image using Probabilistic Hough Transform. A total of 13 features are extracted from each detected line within the image. The features are detailed in Section \ref{Section: Line Methodology}, which includes distance, gradient, and the starting and ending points of the line. The features are labelled as either fractured or non-fractured using a graphical user interface (GUI). The labelled features are used to both train and test the ANN. The architecture of the ANN is described in Section \ref{Section: Line Neural Network Architecture}, in which it consists of four layers, an output and input layer and two hidden layers. PCA is performed on the line features to determine the dominant contributing features that differentiates a fractured line from a non-fractured line. The results of the analysis indicates that the distance in the horizontal direction holds dominance for fractured lines, whilst the distance in the vertical direction holds dominance for non-fractured lines. Further PCA results are detailed in Section \ref{Section: Line Principle Component Analysis}. The line classification of the ANN is detailed in Section \ref{Section: Line-Based Detection Results}, which includes two different experimental set-ups. The first is line-based with image context, in which it evaluates the overall system and the second is line-based without any image context. The second experimental set-up evaluates the performance of the ANN. 

Since the Standard line-based fracture detection approach does not detail all the granule lines found within the fractured area of the X-ray image, a novel ADPO scheme is proposed. This scheme optimizes the parameters of the Probabilistic Hough Transform, such that all detailed lines are detected for further data processing. There are three parameters that are optimized, namely the threshold, minimum line length, and maximum line gap parameters. The optimization and selected values for the three parameters are detailed in Section \ref{Section: Optimized Line-Based Fracture Detection}. The follow-up procedures of the ADPO is similar to the approach of the Standard line-based fracture detection, however the difference is the data that is given to the ANN. Only lines that fall within the leg-bone region of the leg are used to train the ANN, all other lines are disregarded as the lines are classified as knee, foot or flesh. The filtering technique of the bone from flesh lines in the leg area is described in Section \ref{Section: Lines of Interest in the Leg Region}. The performance of the ADPO line-based fracture detection system is detailed in Section \ref{Section: Optimized Line-Based Detection Results}, whereby it uses the same experimental set-ups as the Standard scheme.

\section{Methodology} \label{Section: Line Methodology}

\subsection{Standard Line-Based Fracture Detection} \label{Section: Standard Line-Based Fracture Detection}
\begin{figure}[ht!]
	\begin{center}
		\includegraphics[scale=0.9]{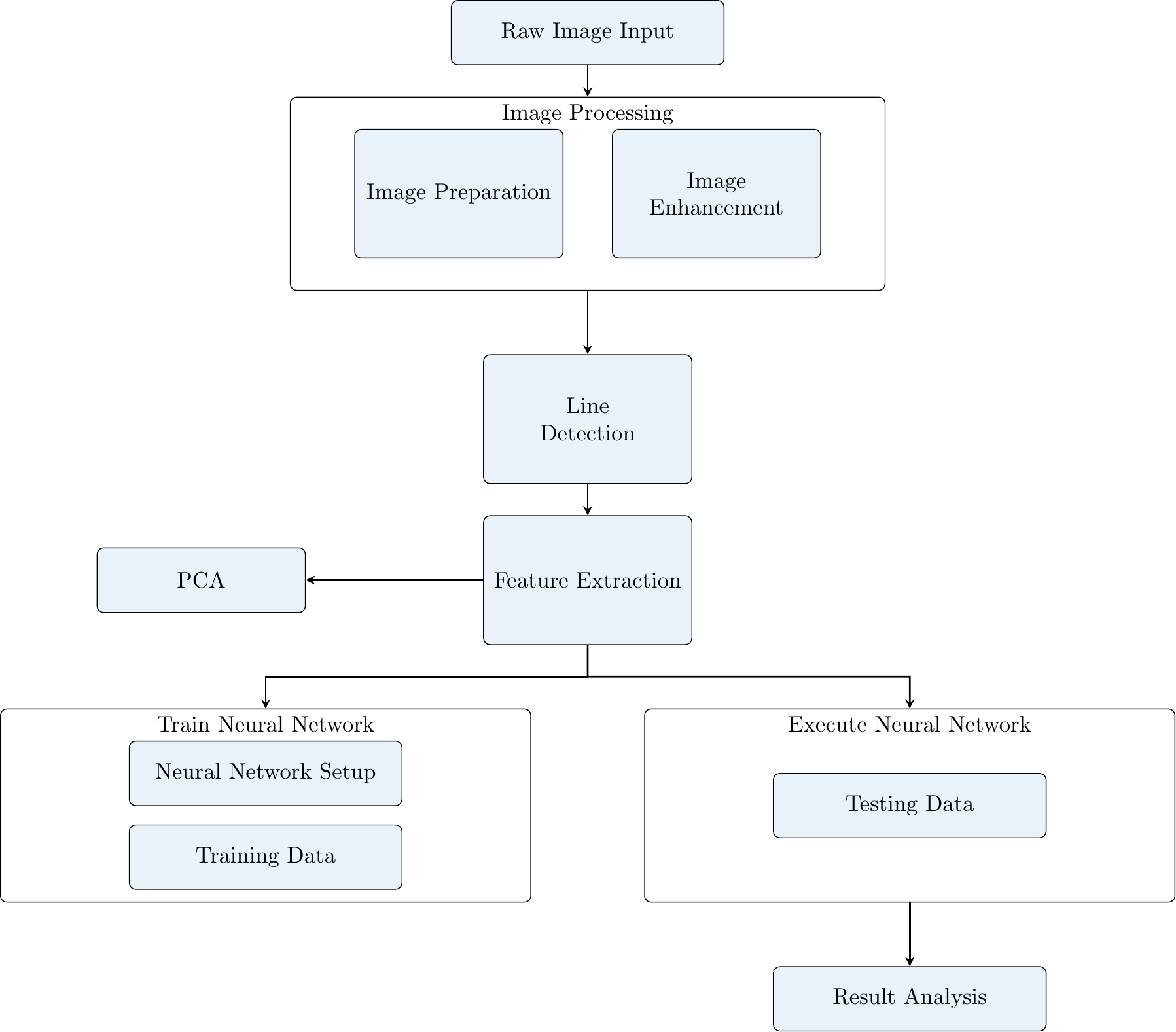}
	\end{center}
	\caption{Flowchart illustrating the procedural flow of the Standard line-based fracture detection}
	\label{fig: line-based system overview}
\end{figure}

The Standard line-based fracture detection follows the procedure in which it first extracts lines from the canny processed image. This is followed by the extraction of 13 features from each line. The features are used to train and test the ANN. Additionally PCA is applied to the features to determine the dominant feature(s) that differentiate fractured and non-fractured lines. The training component of the ANN both sets-up the neural network as well as train it, whilst the execution component is employed for testing the ANN. The results obtained from the evaluation is analysed to assess the ANN's performance. Figure \ref{fig: line-based system overview} illustrates a graphical flow of the Standard line-based fracture detection procedure.

\subsection{Image Enhancement}
All raw X-ray images are converted to greyscale to simplify the image enhancement process, as greyscale images have a single channel compared to RGB where there are three channels. The image enhancement process consists of the removal of white space, pixel equalisation, gamma correction, denoising and unsharp masking. The purpose of the process is to create a high contrast between the long-bone edges and all other pixels within the image. The high contrast ensures that all the image edges are detected by the Canny edge detection operation. The Canny edge detection operation generates a binary image (black and white) with all the long-bone edges highlighted in the image \cite{noauthor_opencv:_nodate}. The Canny edge detection operation is borrowed from the OpenCV2.4 library. The binary images are employed for the line extraction, which lines are extracted based on the edge image objects detected by the Canny edge detection operation. 

\subsection{Line Extraction} \label{Section: Line Extraction}
The line extraction is performed by utilizing Probabilistic Hough Transform. The Probabilistic Hough Transform is a line detection technique in which lines are detected from contrasted image produced by the Canny edge detection technique \cite{stephens1991probabilistic, leavers1993hough}. It uses the Polar system in which a line is expressed in the form shown in \eqref{eq: polar line representation}. A line is defined by rearranging \eqref{eq: polar line representation} to generate \eqref{eq: line point} for point P$(x, y)$. Therefore, each line that passes through $(x_i, y_i)$ is represented by $(r_{\theta}, \theta)$.

\begin{equation}
\label{eq: polar line representation}
y = - \Big(\frac{cos\theta}{sin\theta}\Big)x + \Big(\frac{r_\theta}{sin\theta}\Big)
\end{equation}

\begin{equation}
\label{eq: line point}
r_{\theta} = x cos\theta + y sin\theta
\end{equation}

\begin{figure}[ht!]
	\centering
	\subfigure[Enhanced X-ray image]{\label{fig: enhanced x-ray image}\includegraphics[width=0.49\textwidth]{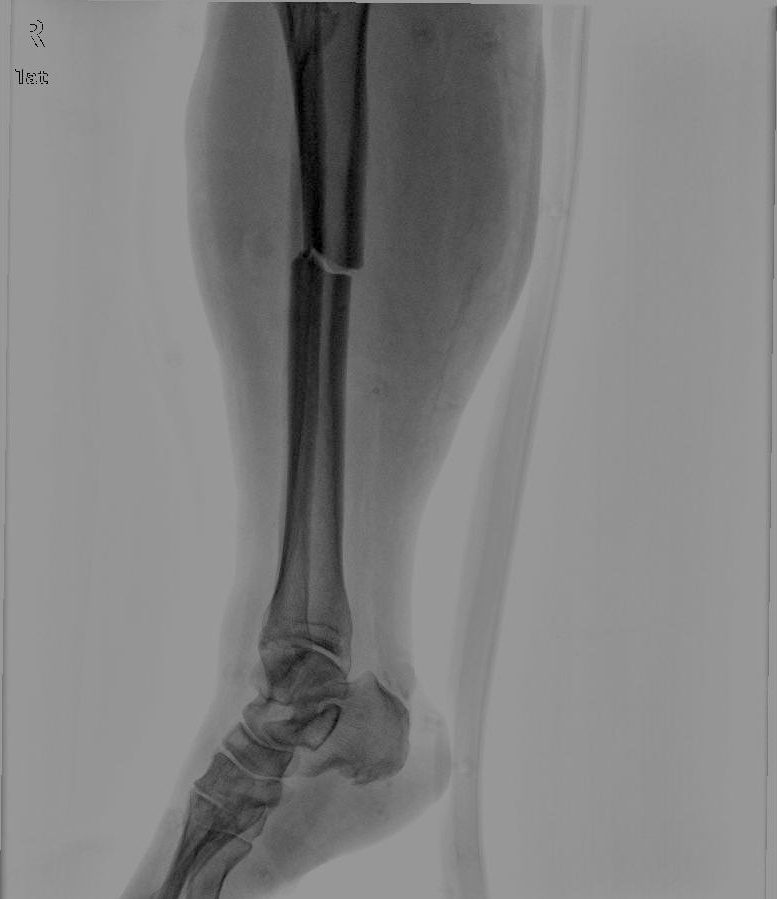}}
	\subfigure[Image of detected lines]{\label{fig: standard line detection}\includegraphics[width=0.49\textwidth]{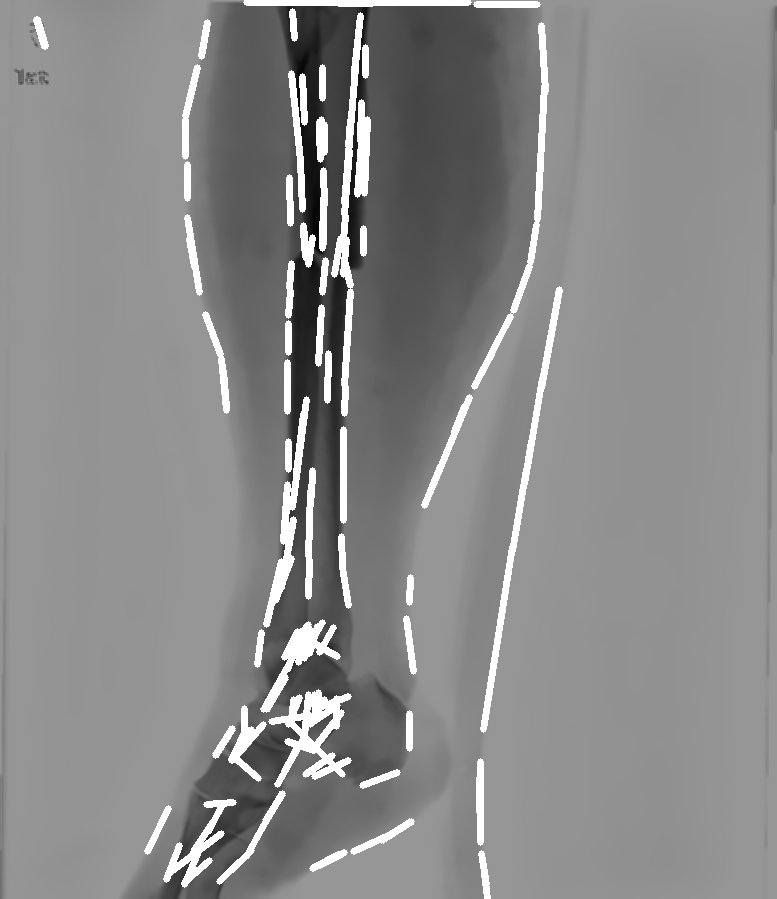}}
	\caption{X-ray images illustrating the detected lines of the Probabilistic Hough Transform with the assigned parameters given in Table \ref{Table: Standard line-based fracture detection Probabilistic Hough Transform parameters}}
	\label{fig: image line extraction}
\end{figure}

The $(r_{\theta}, \theta)$ coordinates are used to detect lines by determining the number of intersections between the curves. An increase in the number of intersections indicates a long line. Therefore, a threshold for the minimum number of intersection is defined for line detection. This is the operation of the Hough Transform, in which it tracks the number of intersections for each $(r_\theta, \theta)$. Detected lines by the Probabilistic Hough Transform are represented in the form of $(x_1, y_1, x_2, y_2)$. The chosen parameter values for the Probabilistic Hough Transform for line detection in the Standard scheme are listed in Table \ref{Table: Standard line-based fracture detection Probabilistic Hough Transform parameters}. The result of the detected lines by the Probabilistic Hough Transform is shown in Figure \ref{fig: standard line detection}.

\vfill
\begin{table}[ht!]
	\centering
	\caption{The detailed Probabilistic Hough Transform parameters for Standard line-based fracture detection approach, along with the reasoning for the selected values.} 
	\label{Table: Standard line-based fracture detection Probabilistic Hough Transform parameters}
	\begin{tabular}{ p{2cm} p{4.5cm} p{2.0cm} p{6cm} }
		\hline
		\textbf{Parameter} & \textbf{Functionality} & \textbf{Assigned Value} & \textbf{Reasoning} \\
		\hline 
		rho, $\rho$ & pixel accumulator for distance resolution of line detection & 1 & The value is assigned to 1 for fine pixel resolution for line detection. \\ \hline
		theta, $\theta$ & angle accumulator for angle resolution in line detection & $\frac{\pi}{180}$ & The angle selected allows for all line angles to be considered for line detection. \\ \hline
		threshold & only accumulators with a value above the threshold are considered as lines & 10 & The threshold value at 10 produces lines with sufficient detailing such that the lines in the fractured region are detected, whilst still maintaining a minimal number of detected lines. \\ \hline
		minimum line length & provides an acceptable minimum line length. Lines that do not meet this requirement are rejected & 25 & The chosen value assists with eliminating lines that are too short. The short lines are product of the remaining noise from the enhanced image. \\ \hline
		maximum line gap & defines the maximum line gap between points that link the same line & 10 & The value assigned to the parameter is chosen based on the generated lines, whereby the lines generated cover all critical information within the image but does not over extend the generated lines. An increased value, generates extended lines which implies that there is existing information that is not present in the image, whereas a decreased value shortens the lines and misinterprets the image information. \\ \hline
	\end{tabular}
\end{table}

\subsection{Line-based Feature Extraction} \label{Section: Line-based Feature Extraction}
A set of 13 features are extracted from each detected line. The features are a summarised representation of the lines detected from image edge objects found in the X-ray image. These features are used as inputs into the ANN. The purpose of extracting the features is to quantify and provide crucial information about the lines to the ANN that differentiates a fractured line from a non-fractured line. The quantification of the features reduces the complexity of the ANN in both training and execution. The features are extracted based on the starting and ending points in the form of $L(x_1, y_1, x_2, y_2)$, provided by the Probabilistic Hough Transform. These extracted features are listed and detailed in Table \ref{Table: Line Extracted Features}.

\begin{table}[ht!]
	\centering
	\caption{The details of the extracted line features along with the feature notation and extraction methodology}\label{Table: Line Extracted Features}
	\begin{tabular}{ c p{2.5cm} p{1.8cm} p{1.5cm} p{7cm} }
		\hline
		& \textbf{Extracted Feature} & \textbf{Notation} & \textbf{Abv.} & \textbf{Extraction Methodology} \\
		\hline 
		1 & x start & $x_1$ & X1 & $x_1$ is the x-value of the starting point \\ \hline
		2 & y start & $y_1$ & Y1 & $y_1$ is the y-value of the starting point \\ \hline
		3 & x end & $x_2$ & X2 & $x_2$ is the x-value of the ending point, where $x_2 > x_1$ \\ \hline
		4 & y end & $y_2$ & Y2 & $y_2$ is the y-value of the ending point \\ \hline
		5 & distance & $d$ & DIST & The \textit{distance} feature is extracted using $x_1$, $y_1$, $x_2$, and $y_2$. The distance calculation is expressed in \eqref{eq: line distance equation}. 
		\begin{equation}
		\label{eq: line distance equation}
		d = \sqrt{(x_2-x_1)^2+(y_2-y_1)^2}
		\end{equation}  \\ \hline
		6 & gradient & $\theta_\text{L}$ & G & The gradient feature is determined using $x_1$, $y_1$, $x_2$, and $y_2$. It is determined using \eqref{eq: line gradient equation}.
		\begin{equation}
		\label{eq: line gradient equation}
		\theta_\text{L} = tan^{-1} \Big(\frac{x_2 - x_1}{y_2 - y_1} \Big)
		\end{equation} \vspace{-5mm} \\ \hline
		7 & x-Midpoint & $x_\text{M}$& X-MID & The x-Midpoint feature is determined using $x_1$ and $x_2$. The midpoint in the x-direction is calculated using \eqref{eq: line x-Midpoint}.
		\begin{equation}
		\label{eq: line x-Midpoint}
		x_\text{M} = \frac{x_1 + x_2}{2}
		\end{equation} \vspace{-5mm} \\ \hline
		8 & y-Midpoint & $y_\text{M}$ & Y-MID & The y-Midpoint feature is calculated using $y_1$ and $y_2$. The midpoint in the y-direction is calculated using \eqref{eq: line y-Midpoint}.
		\begin{equation}
		\label{eq: line y-Midpoint}
		y_\text{M} = \frac{y_1 + y_2}{2}
		\end{equation} \vspace{-5mm} \\ \hline
		9 & x-Difference & $\Delta x$ & X-DIFF & The x-Difference feature is determined using $x_1$ and $x_2$. It determines the difference between the x-values.
		\begin{equation}
		\label{eq: x-difference}
		\Delta x = x_2 - x_1
		\end{equation} \vspace{-5mm} \\ \hline	
		10 & y-Difference & $\Delta y$ & Y-DIFF & The y-Difference feature is determined using $y_1$ and $y_2$, in which it determines the difference between the y-values.
		\begin{equation}
		\label{eq: y-difference}
		\Delta y = y_2 - y_1
		\end{equation} \vspace{-5mm} \\ \hline	
		11 & x-distance & $d_\text{x}$ & X-DIST & This feature is derived from the Pythagoras theorem using \eqref{eq: x-distance}, but the distance in the x-direction. 
		\begin{equation}
		\label{eq: x-distance}
		d_\text{x} = d cos(\theta_\text{L})
		\end{equation} \vspace{-5mm} \\ \hline	
		12 & y-distance & $d_\text{y}$ & Y-DIST & The y-distance feature similar to the x-distance feature, however it is given in the y-direction. The y-direction is extracted using \eqref{eq: y-distance}.
		\begin{equation}
		\label{eq: y-distance}
		d_\text{y} = d sin(\theta_\text{L})
		\end{equation} \vspace{-5mm} \\ \hline
		13 & gradient deviation & $\Delta \theta$ & G-DEV & The gradient deviation feature indicates the amount that the current gradient deviates from the most frequently occurred gradient, $\theta_\text{Ref}$. The gradient deviation is obtained using \eqref{eq: gradient deviation}.
		\begin{equation}
		\label{eq: gradient deviation}
		\Delta\theta = \mid \theta_{\text{Ref}} - \theta_\text{L}\mid
		\end{equation} \vspace{-5mm} \\ \hline
	\end{tabular}
\end{table}

\newpage
\subsection{Feature Correlation Analysis}
The Pearson product-moment correlation coefficient is used to measure the dependency between each feature extracted from 39,224 lines. The correlation coefficient, $c$ is calculated using \eqref{eq: correlation coefficient}. The correlation coefficient is an indication of the strength of the association between two features. It is given between a range of -1 to 1, which quantifies both strength and direction of the association between two features. The value ``$-1$" indicates a strong negative relation. This means that as one feature increases in value, the other decreases. A ``$0$" value shows that there is no association, and the "$1$" indicates a strong positive relation between two features, as either feature value increases, the other increases as well.

\begin{equation}
\label{eq: correlation coefficient}
c = \frac{cov(\boldsymbol{x},\boldsymbol{y})}{\sqrt{s_x^2 \cdot s_y^2}},
\end{equation}
where $cov(\boldsymbol{x},\boldsymbol{y})$ is the covariance of $\boldsymbol{x}$ and $\boldsymbol{y}$ and $s_x^2$ and $s_y^2$ are the sample variances of $\boldsymbol{x}$ and $\boldsymbol{y}$. The covariance and samples are defined as follows:

\begin{align*}
cov(\boldsymbol{x}, \boldsymbol{y}) &= \frac{\sum(x-\bar{x})(y-\bar{y})}{n - 1} \\
s^2_x &= \frac{\sum(x-\bar{x})^2}{n - 1}\\
s^2_y &= \frac{\sum(x-\bar{x})^2}{n - 1}
\end{align*}

\begin{figure}[ht!]
	\centering
	\includegraphics[scale=0.55]{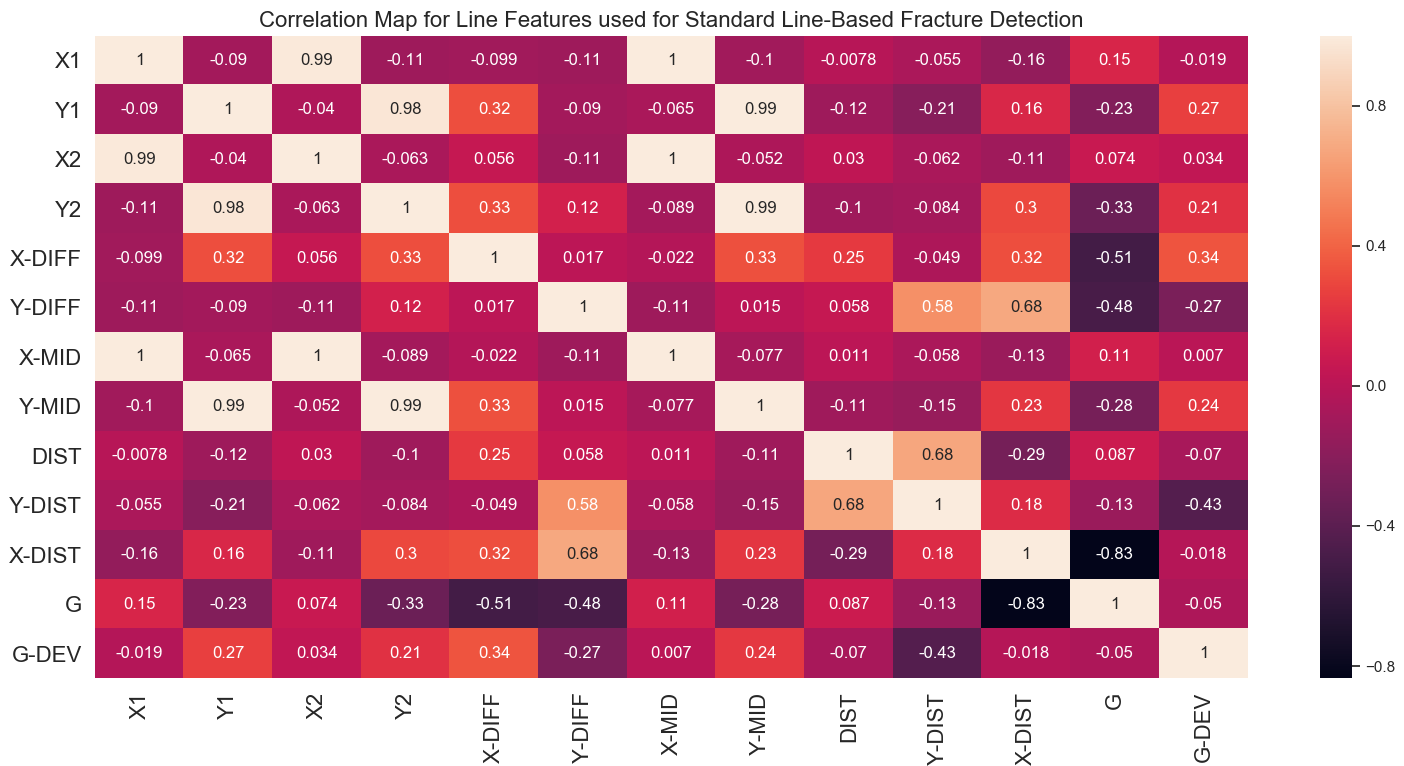}
	\caption{Correlation Map illustrating the association between each of the 13 features extracted}
	\label{fig: line_correlation_map}
\end{figure}

The purpose of a correlation map is to provide a visual association between each extracted line feature. The association provides an indication of the features that are dependent on one another, as well as features that are independent from all other features. The correlation map for the 13 extracted line features is illustrated in Figure \ref{fig: line_correlation_map}. The correlation map indicates that the \textit{gradient deviation} and \textit{x-Difference} feature are less reliant on all other features and are more independent as they their relations with the other features are either weak positive or weak negative. Independent features are crucial as they have minimal redundant information. This allows for some discrepancy in the input information used to train the ANN. Redundant features contain duplicated information and as a result, they do not provide enough information to allow for the differentiation between fracture lines and non-fractured lines. The correlation map presented in Figure \ref{fig: line_correlation_map} contains a minimal amount of features that have a strong positive (light blocks) or strong negative (dark blocks) association with other features.

\subsection{Principal Component Analysis} \label{Section: Line Principle Component Analysis}
Principal Component Analysis (PCA) is a linear technique that performs dimensionality reduction through the process of embedding the data into a linear subspace of low-dimensionality. The low-dimensional representation describes the variance found within the data \cite{Jolliffe2016}. The purpose of utilizing PCA in this paper is to identify the dominant line features to determine the main contributing factors that distinguishes whether a line is considered fractured or non-fractured. PCA operates in such a manner that the feature with the most variation is considered a dominant feature. The variation of the features is one of the contributing factors for the classification of fractured and non-fractured lines. The variation of a feature indicates that the feature values varies for different line characteristics. PCA is applied to a number of lines, $m$, where $m = 39,224$ for a number of features, $n$ where $n = 13$. A matrix, $X$ with the dimensions of $m \times n$ is created to represent each feature of each line. The matrix is normalised to generate matrix, $X'$ in which PCA is applied to. PCA produces a $n \times n$ covariance matrix, $A$. The eigenvalues, $\boldsymbol{\lambda}$ and eigenvectors, $\boldsymbol{v}$ are generated using the covariance matrix. The eigenvalue, $\lambda_i$ is a scaler which provides the magnitude of the varying values for each $i$-th feature, where $i \in \{1, 2, 3, ..., n\}$. The higher the scaler value, the more variation the feature contains, whereas a lower scaler value shows that there is little to no variation within the feature. However, it is difficult to assign the appropriate eigenvalue to its associated feature as the eigenvalues are sorted from largest to smallest due to the PCA procedure. Therefore, the eigenvectors are utilized, as each element in the eigenvector is ordered in the same manner as the columns of the matrix, $X$. Thus, the $i$-th feature is represented by element, $e_{ij}$ in eigenvector, $\boldsymbol{e}_j$. The determination of the contribution for each feature is entailed as follows:

\begin{enumerate}
	\item Sum all elements in eigenvector, $\boldsymbol{e}_j$ to generate a vector of summed elements, $\boldsymbol{s} = \{s_1, s_2, s_3, s_j, ..., s_n\}$, where $n$ is the number of eigenvectors. The calculation of $s_j$ for the $j$-th eigenvector is expressed in \eqref{eq: eigenvector sum}.
	
	\begin{equation}
	\label{eq: eigenvector sum}
	s_j = \sum_{i=1}^{n} e_{ij}
	\end{equation}
	\item Utilizing $\boldsymbol{s}$, convert each element in the eigenvectors to a ratio, $r_{ij}$. The eigenvectors are arranged in such a manner that it produces matrix, $R$ with elements $r_{ij}$. Each element, $r_{ij}$ is calculated using \eqref{eq: PCA element percentage}.
	\begin{equation}
	\label{eq: PCA element percentage}
	r_{ij} = \frac{e_{ij}}{s_j} 
	\end{equation} 
	\item The overall feature contribution, $c_{i}$ obtained by averaging all elements in the $i$-th row of matrix, $R$. This calculation is expressed in \eqref{eq: feature contribution}. The feature contribution is associated to the $n$ extracted features from the lines.
	\begin{equation}
	\label{eq: feature contribution}
	c_j = \frac{\sum_{j = 1}^{n} r_{ij}}{n} 
	\end{equation}
\end{enumerate}

\begin{figure}[ht!]
	\centering
	\includegraphics[scale=0.65]{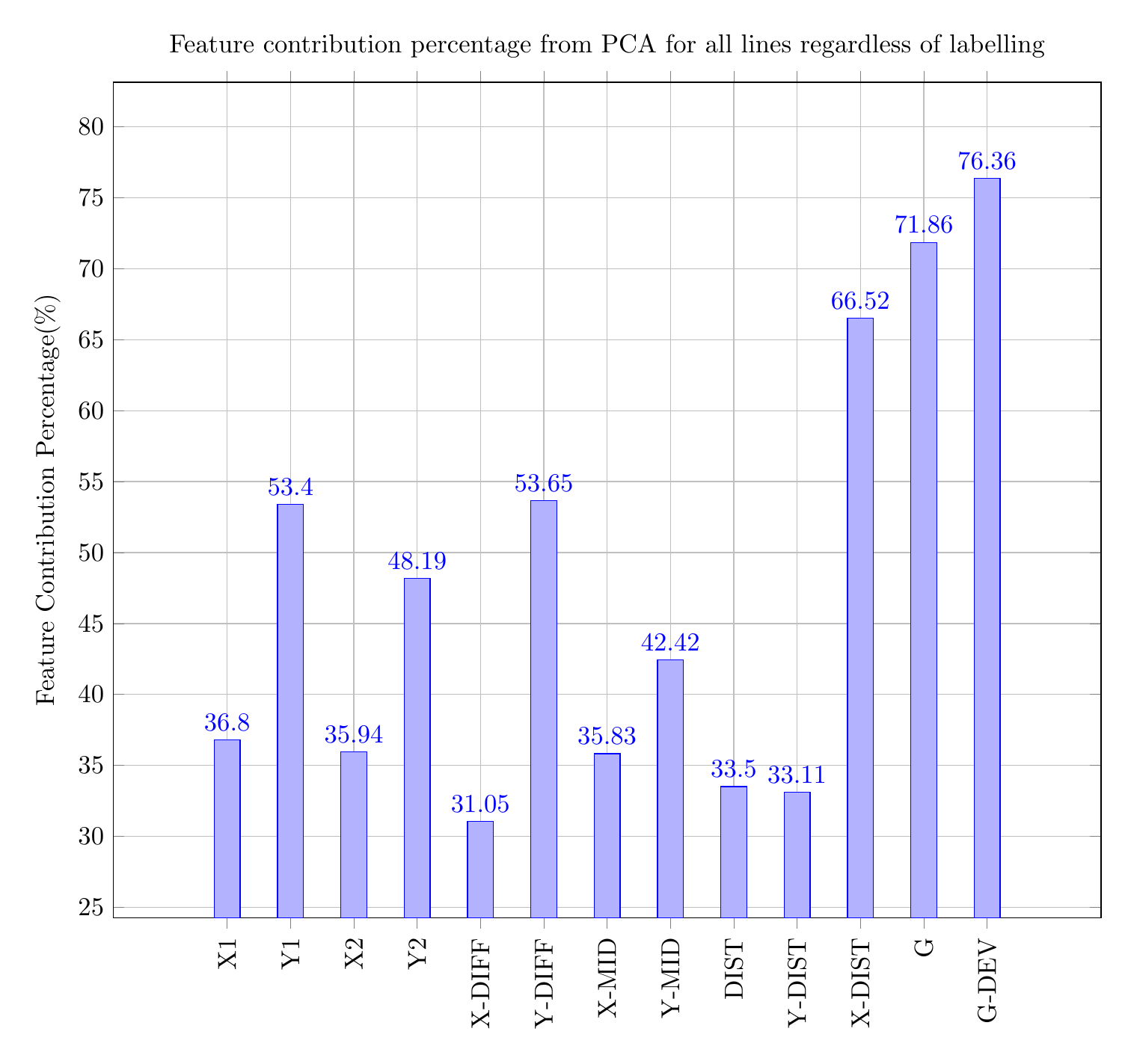}
	\caption{Histogram illustrating the results of the feature contribution from PCA for all extracted lines regardless of line classification}
	\label{fig: All line feature pca}
\end{figure}

\begin{figure}[ht!]
	\centering
	\includegraphics[scale=0.65]{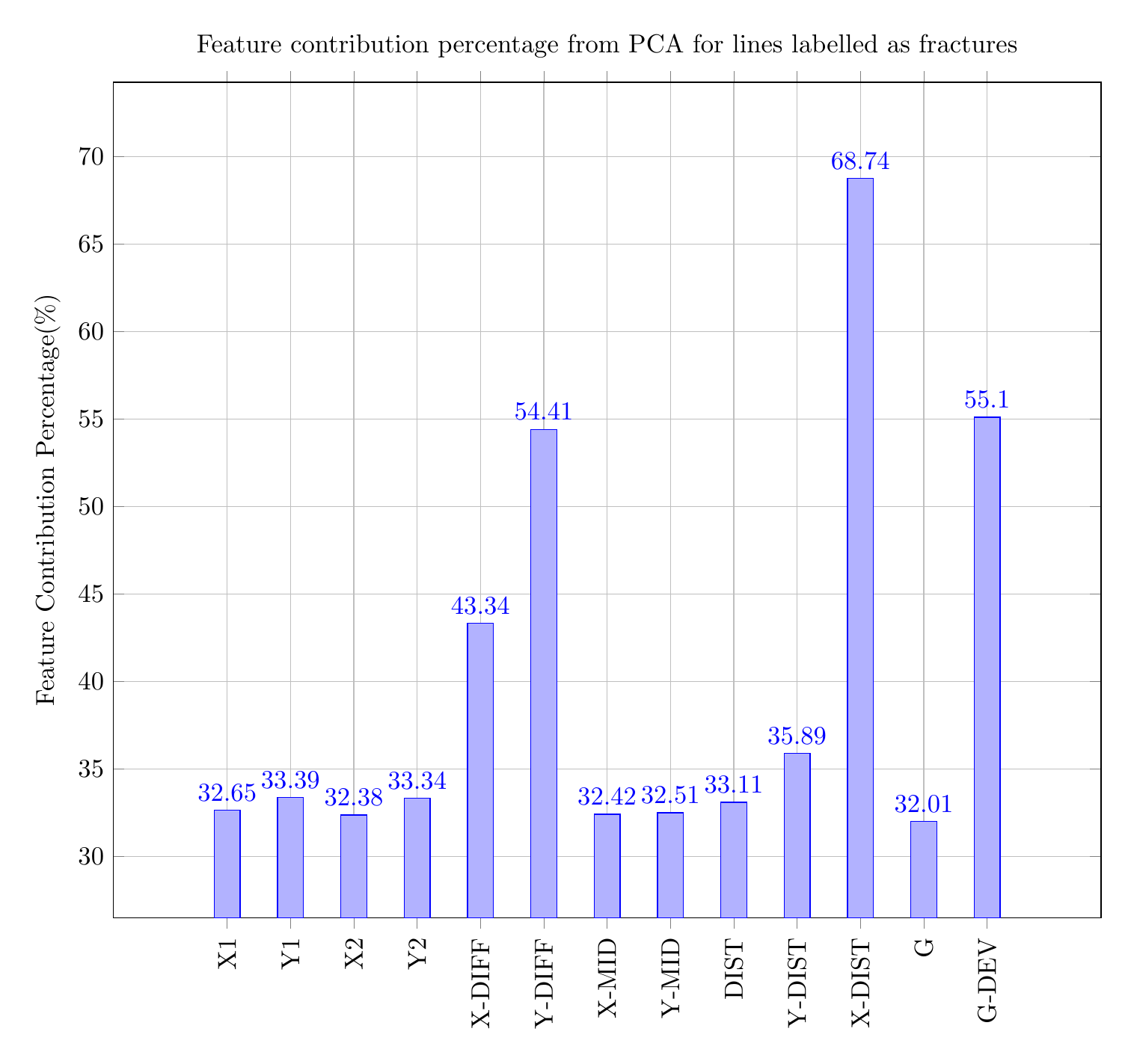}
	\caption{Histogram illustrating the results of the feature contribution from PCA for extracted lines labelled as fractured}
	\label{fig: line pca fracture}
\end{figure}

\begin{figure}[ht!]
	\begin{center}
		\includegraphics[scale=0.65]{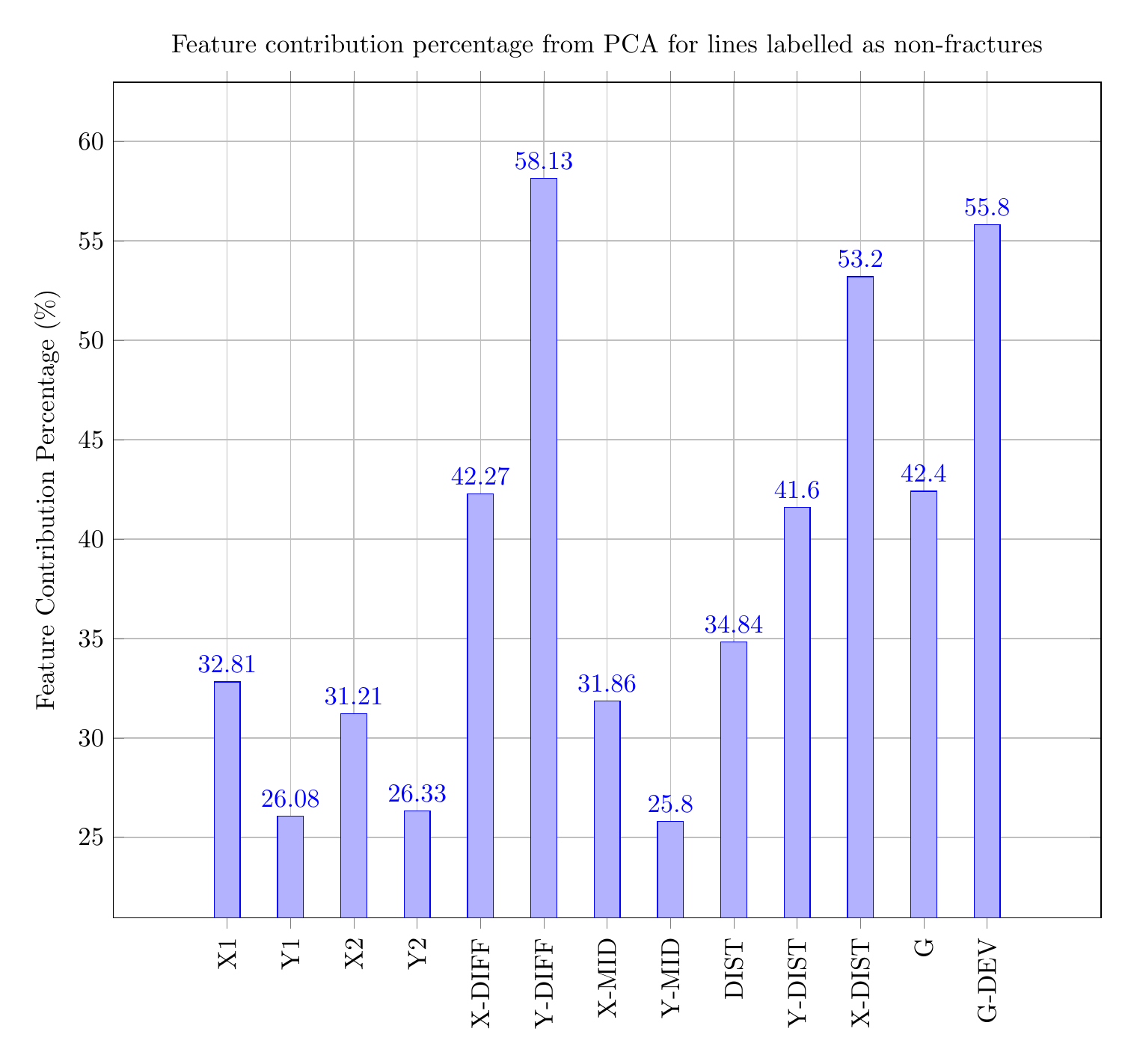}
	\end{center}
	\caption{Histogram illustrating the results of the feature contribution from PCA for extracted lines labelled as non-fractured}
	\label{fig: line pca non-fracture}
\end{figure}

Figures \ref{fig: All line feature pca}, \ref{fig: line pca fracture}, and \ref{fig: line pca non-fracture} illustrates the results of the PCA feature contribution. These results are crucial as it shows the indicative feature(s) that contribute most to the differentiation of a fractured and non-fractured line. Additionally, it is an indication of the feature that holds the most information about the line characteristics. Figure \ref{fig: All line feature pca} shows the feature contribution for 39,224 extracted lines from 53 images. Figure \ref{fig: All line feature pca} disregards the labelling of the lines. The most dominant feature in Figure \ref{fig: All line feature pca} is \textit{gradient deviation} with 76.36\% contribution, followed by the \textit{gradient} and \textit{x-distance} feature. Therefore, despite the labelling of the lines, the \textit{gradient deviation}, \textit{gradient} and \textit{x-distance} has the most varying values from the extracted lines. Thus, the three features holds crucial information about the line characteristics.

Figure \ref{fig: line pca fracture} presents the feature contribution for 15,561 extracted line features that are labelled as fractures. The results show that there are three very distinct features that are dominant for fractured lines, namely, \textit{x-distance}, \textit{gradient deviation} and \textit{y-difference}. The feature \textit{x-distance} has the most varying values for fractured lines. This is valid as the orientation of fractured lines are generally positioned in a horizontal direction. Additionally, the feature \textit{y-difference} is valid, because the fractured lines vary greatly in vertical length. For the \textit{x-difference} feature, this indicates that the fractured lines vary greatly in length in the horizontal direction.

Figure \ref{fig: line pca non-fracture} shows the three main dominant contributing features for non-fractured lines, which are \textit{x-difference}, \textit{gradient deviation}, and \textit{x-distance}. The results indicate that the fractured and non-fractured features share two dominant features, namely \textit{gradient deviation} and \textit{y-difference}. This can cause confusion to the ANN in the classification of fractured and non-fractured lines, however the features differ in contribution for fractured and non-fractured lines. The results further shows that there are minor dominant features for non-fractured lines, namely, \textit{x-difference}, \textit{y-distance}, and \textit{gradient} to assist with the differentiation between fractured and non-fractured lines. 

The application of the PCA provides sufficient detailing about the contributions of each feature for the both fractured and non-fractured lines, as well as all lines in general. There are features that have minimal contribution to the line information in which it can be eliminated to reduce the training and execution complexity of the ANN. However, with the potential of confusion between the fractured and non-fractured lines all extracted features are considered given that there are only 13 features providing crucial information about the extracted line.

\subsection{Neural Network Structure} \label{Section: Line Neural Network Architecture}
The architecture of the ANN for the Standard line-based fracture detection scheme consists of four layers: one input and output layer, and two hidden layers. The number of nodes, $n$ in the input layer is 16 nodes ($n = 16$). There are three additional nodes compared to the number of extracted line features in the ANN input layer. The three additional nodes are for the labelling of the X-ray image regions, namely, the knee, leg and foot region. Thus, if the line belongs within the leg region, the value ``1" is assigned to the associated leg feature indicator, whilst a ``0" value is assigned to both knee and foot feature indicator. The assignment of the knee, leg and foot region is performed during line feature extraction. Therefore, the knee, leg and foot region assignments are considered as additional features to the extracted line features. 

There are two hidden layers in the architecture of the ANN. The number of layers within the neural network defines the ANN complexity \cite{hornik1991approximation}. An increased number of hidden layers increases training complexity of the ANN, since it will increase the number of iterations until the desired error is obtained \cite{cybenko1989approximation}. Furthermore, increasing the number of hidden layers does not yield a higher accuracy from the ANN. However for a single hidden layer, the ANN is over simplified and is prone to over-fitting. Therefore, increasing the difficulty of obtaining a high accuracy. This leads to the selection of having two hidden layers for the architecture of the ANN for Standard line-based fracture detection scheme. Each hidden layer has $n+1$ nodes. The purpose of the additional node in the hidden layer compared to the input layer is to introduce an additional vote of input before reaching the output layer \cite{hornik1991approximation, hinton2006fast}.

The output layer only has one node, the outcome of the output node is expressed as $O_f$, where $-1 \leq O_{f} \leq 1$. The range between ``-1" to ``1" is defined by the labelling of the training data, whereby fractured lines are assigned to ``1" as its target output and non-fractured lines are assigned to ``-1". The details of the data labelling process is discussed in Section \ref{Section: Data Labelling}. The final detection outcome, $O$ classifies the input line as fractured or non-fracture with ``true" or ``false. Thus, $O \in \{true, false\}$ and is expressed in \eqref{eq: nueral network outcome definition}, where $true$ defines a fracture and $false$ defines a non-fracture. The ANN training and network set-up is detailed in Table \ref{Table: line neural network detail set-up}.

\begin{equation}
\label{eq: nueral network outcome definition}
O = \left\{
\begin{array}{ll}
true, & 0 \leq O_{f} \leq 1 \\
false, & -1 \leq O_{f} < 0\\
\end{array}
\right.
\end{equation}

\begin{table}[ht!]
	\centering
	\caption{The ANN training set-up for the Standard line-based fracture detection approach}
	\label{Table: line neural network detail set-up}
	\begin{tabular}{p{4.0cm} p{7cm} p{3.0cm} }
		\hline
		\textbf{ANN Detail Set-Up} & \textbf{Functionality} & \textbf{Assigned Value} \\ 
		\hline
		No. of ANN epochs & The number of epochs defines the maximum number of iterations allowed during the training of the ANN. If the error of the ANN does not meet the desired error value, the network continues training until the number of iterations has reached the defined epochs value & 50,000 epochs. \\ \hline 
		Desired Error & The desired error value is the minimal error allowed for the ANN during the training, as weights cannot perfectly match the input to out value , whilst still maintaining generality over large data sets. & 0.0001 \\ \hline	
		Training Data Order & The data given to the ANN for training is potentially ordered. An ordered training data set can swing the ANN weights from one decision to another. Thus, the training data is shuffled such that there is no favouritism is given over a selected set of data. & ``shuffled" \\ \hline
	\end{tabular}
\end{table}

\begin{figure}[ht!]
	\begin{center}
		\includegraphics[scale=0.9]{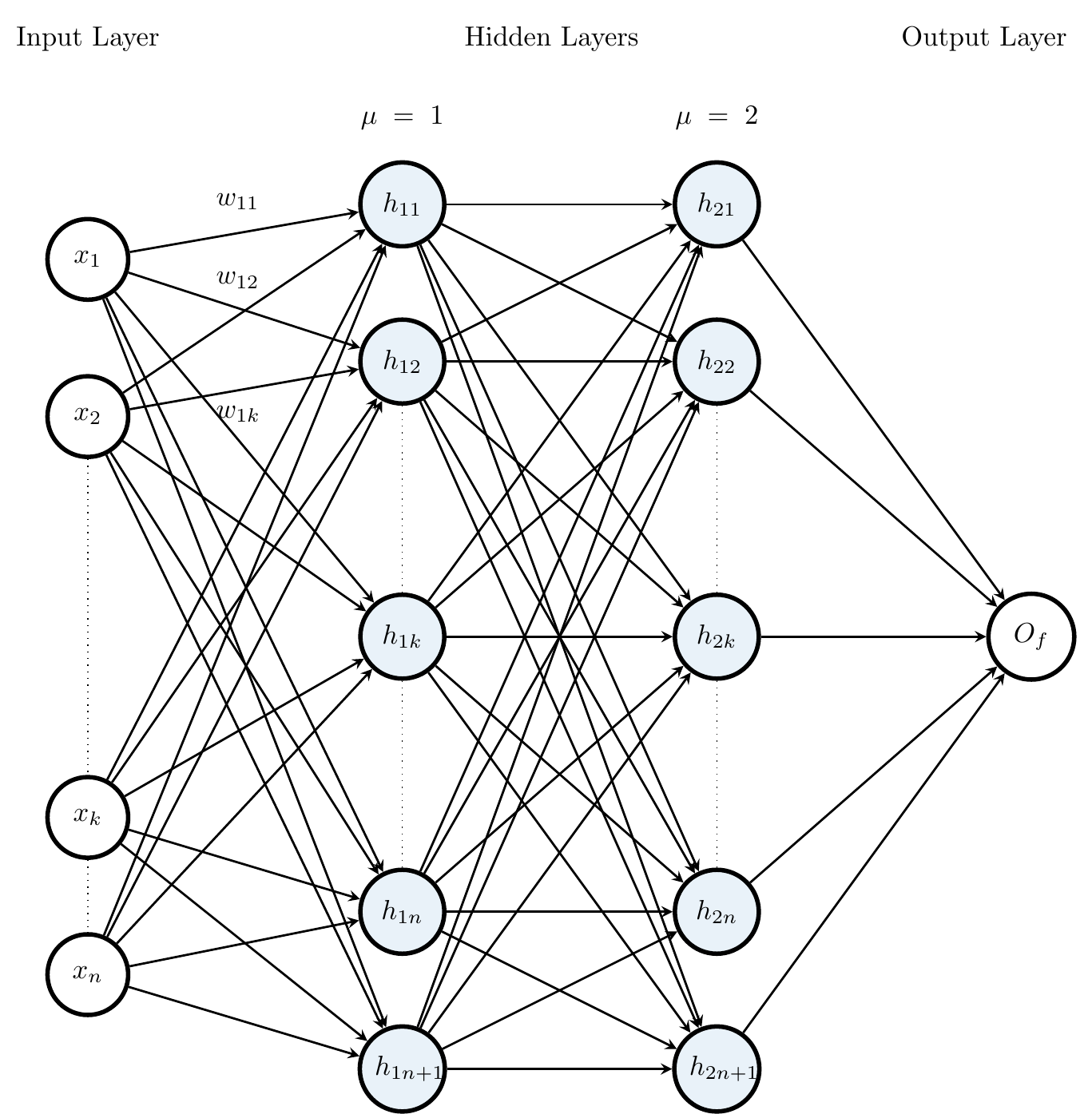}
	\end{center}
	\caption{Diagram showing the artificial neural network architecture for Standard line-based fracture detection scheme, where $\mu$ is indicative of the hidden layers.}
	\label{fig: line neural netwrok setup}
\end{figure}

\newpage
\subsection{Data Line Labelling} \label{Section: Data Labelling}
\begin{figure}[ht!]
	\centering
	\subfigure[Fracture Line Selection]{\label{fig: GUI line presentation}\includegraphics[width=0.49\textwidth]{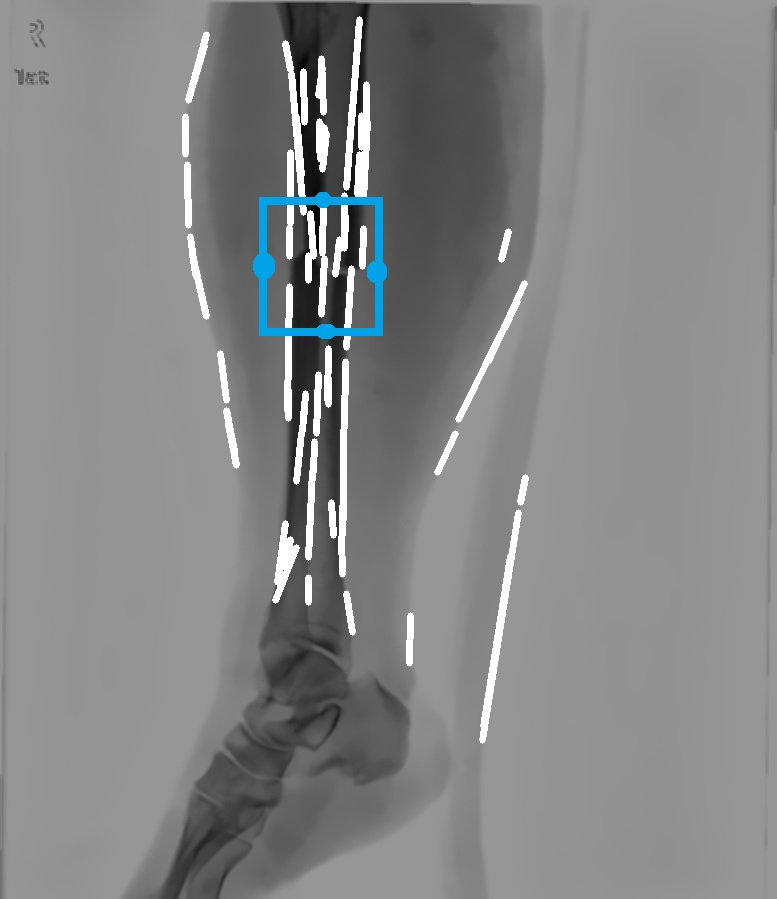}}
	\subfigure[Highlighted Fracture Lines]{\label{fig: RawImagebfg(3)}\includegraphics[width=0.49\textwidth]{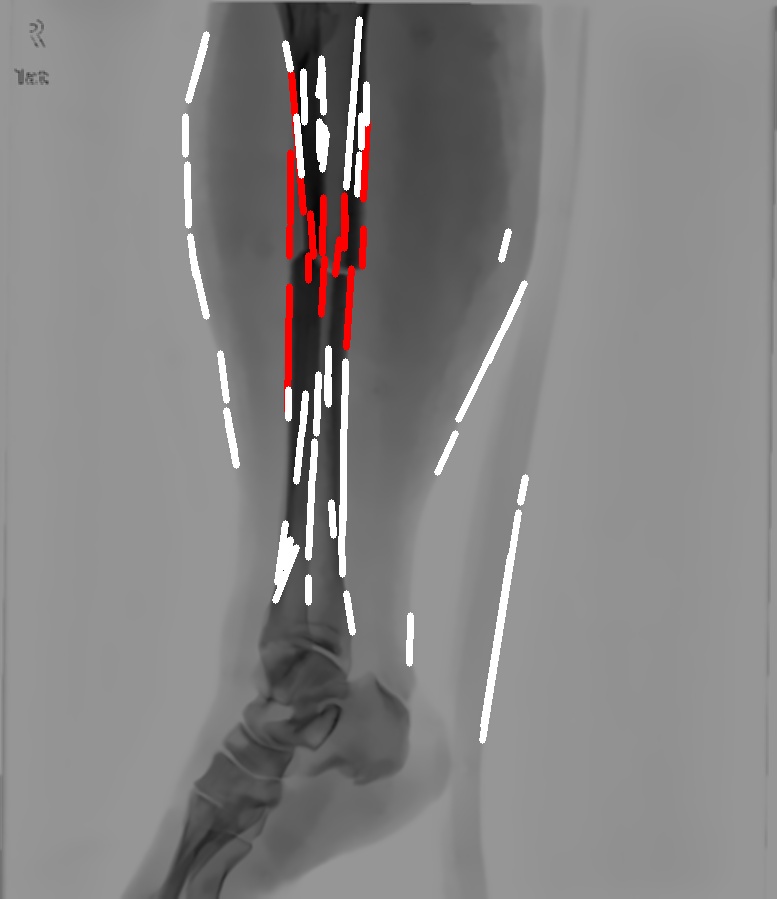}}
	\caption{Images illustrating the labelling of the detected lines} 
	\label{fig: labelling extracted lines}
\end{figure}

The data labelling is performed using a graphical user interface (GUI). The data is labelled visually through the GUI by a user. The extracted lines from the Probabilistic Hough Transform is drawn onto the enhanced image of the original X-ray image, shown in Figure \ref{fig: labelling extracted lines}. Before the lines are labelled, further processing is performed to isolate the lines within the knee and foot region from the lines in the leg region. Thus, the labelling process is restricted to the lines in the leg region. 

The labelling process is performed manually by a human professional, whereby the user selects the region of the fracture. The lines in the selected region are labelled as fractures while the remaining lines are labelled as non-fractures. The regional labelling approach is chosen, as it mimics the visual detection approach that a medical professional utilises. This approach is chosen over the labelling of individual lines, as it is far too time-consuming and impractical. The impracticality of labelling individual lines stems from the visual information that a single line provides. A single line provides little to no evidence of whether it is fractured or non-fractured. Whereas, a group of lines provides more context about the line classification. The regional data labelling approach, provides adequate information about the lines based on its location and neighbouring lines. A line is considered a fracture if either its starting or ending point is within the selected area. However, this approach is not perfect, as it mislabels lines that are non-fractures but fall within the selected region as fractures.

\section{Results of the Standard Line-Based Fracture Detection} \label{Section: Line-Based Detection Results}
\subsection{Artificial Neural Network Experimental Set-Up} \label{Section: ANN Experiement set-up}
The ANN is evaluated based on its performance to make accurate fracture and non-fracture classification. The set-up of the evaluation consists of both system and ANN evaluation. For the system evaluation, a total of 20 images are used to evaluate the system. There are 20 individual image cases for the system evaluation. Each case trains the ANN with a number of images ranging from 1 to 20. This evaluation provides the ANN with lines that has context about the image it is extracted from. The images, $i$, are randomly selected from a set of training images,$n$. Hence, $i \in \{1, 2, 3, ..., n\}$, where $n = 29$. There are an average of 740 lines per image, and the average ratio of fractures to non-fractures is $1:1.52$. Each case has 10 simulations. This is to obtain an average accuracy for the system as each simulation accuracy outcome differs slightly from one another. The slight differences are a result of the randomly initialised weights in the ANN as well as the randomly selected images used for each training case. 23 images are used to test the system for each case to ensure that each case and simulation is evaluated fairly.

\subsection{System and Artificial Neural Network Results}
The results of the system evaluation is presented in Table \ref{Table: Line Neural Network Accuracy}, whereby the minimum, average and maximum accuracies for each case is detailed. The system was evaluated using a total of 11,910 lines that are extracted from 23 images. From the 11,910 lines, 4,707 lines are fractured and 7,203 are non-fractured lines.

\begin{table}[ht!]
	\centering
	\caption{The results for the system's minimum, average and maximum accuracy of the system for 20 cases over 10 simulations for the Standard line-based fracture detection scheme}
	\label{Table: Line Neural Network Accuracy}
	\begin{tabular}{c c c c }
		\hline
		\textbf{No. Trained Images, $c_i$} & \textbf{Min Accuracy (\%)} & \textbf{Average Accuracy (\%)} & \textbf{Max Accuracy (\%)} \\
		\hline 
		1 & 68.606 & 72.764 & 75.777 \\
		2 & 60.462 & 69.9842 & 75.206 \\
		3 & 60.831 & 71.0134 & 75.113 \\
		4 & 66.255 & 71.8111 & 75.718 \\
		5 & 60.336 & 70.1906 & 75.407 \\
		6 & 66.314 & 72.1956 & 75.466 \\
		7 & 69.639 & 72.2939 & 75.743 \\
		8 & 60.571 & 71.3233 & 75.919 \\
		9 & 67.674 & 71.7062 & 75.743 \\
		10 & 68.85 & 72.7254 & 75.449 \\
		11 & 67.674 & 72.5499 & 75.76 \\
		12 & 67.599 & 71.5242 & 75.332 \\
		13 & 69.018 & 72.4467 & 74.845 \\
		14 & 72.704 & 74.2133 & 75.651 \\
		15 & 60.579 & 69.7346 & 75.374 \\
		16 & 60.487 & 70.0982 & 74.433 \\
		17 & 67.389 & 72.6766 & 75.306 \\
		18 & 61.226 & 69.519 & 75.928 \\
		19 & 60.437 & 68.9798 & 75.55 \\
		20 & 70.848 & 73.7094 & 75.197 \\
		\hline
	\end{tabular}
\end{table}

\begin{figure}[ht!]
	\centering
	\includegraphics[scale=0.8]{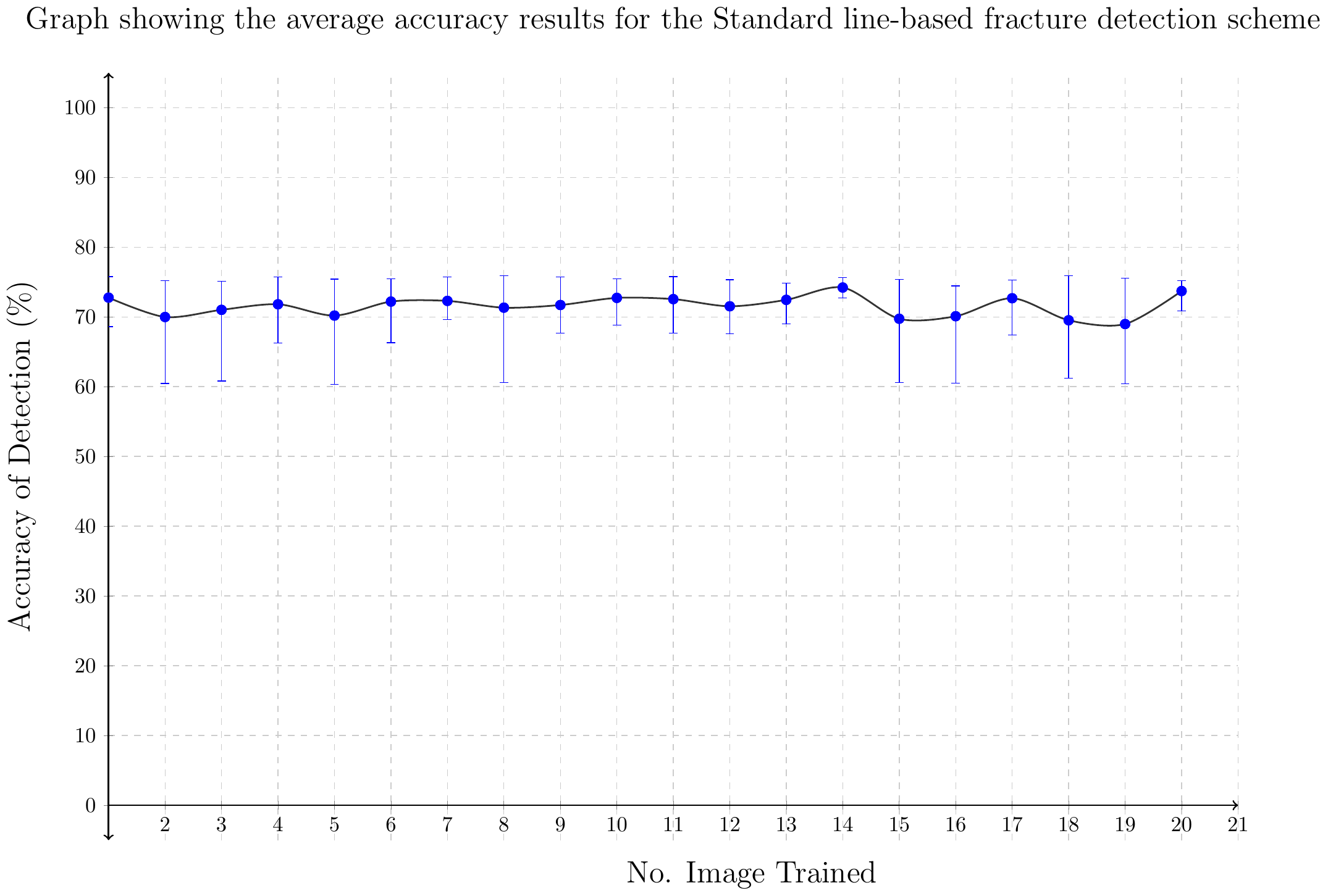}
	\caption{Graph illustrating the average accuracy for 20 cases over 10 simulations for the Standard line-based fracture detection scheme}
	\label{fig: line accuracy graph with error bars}
\end{figure}

\begin{figure}[ht!]
	\centering
	\includegraphics[scale=0.8]{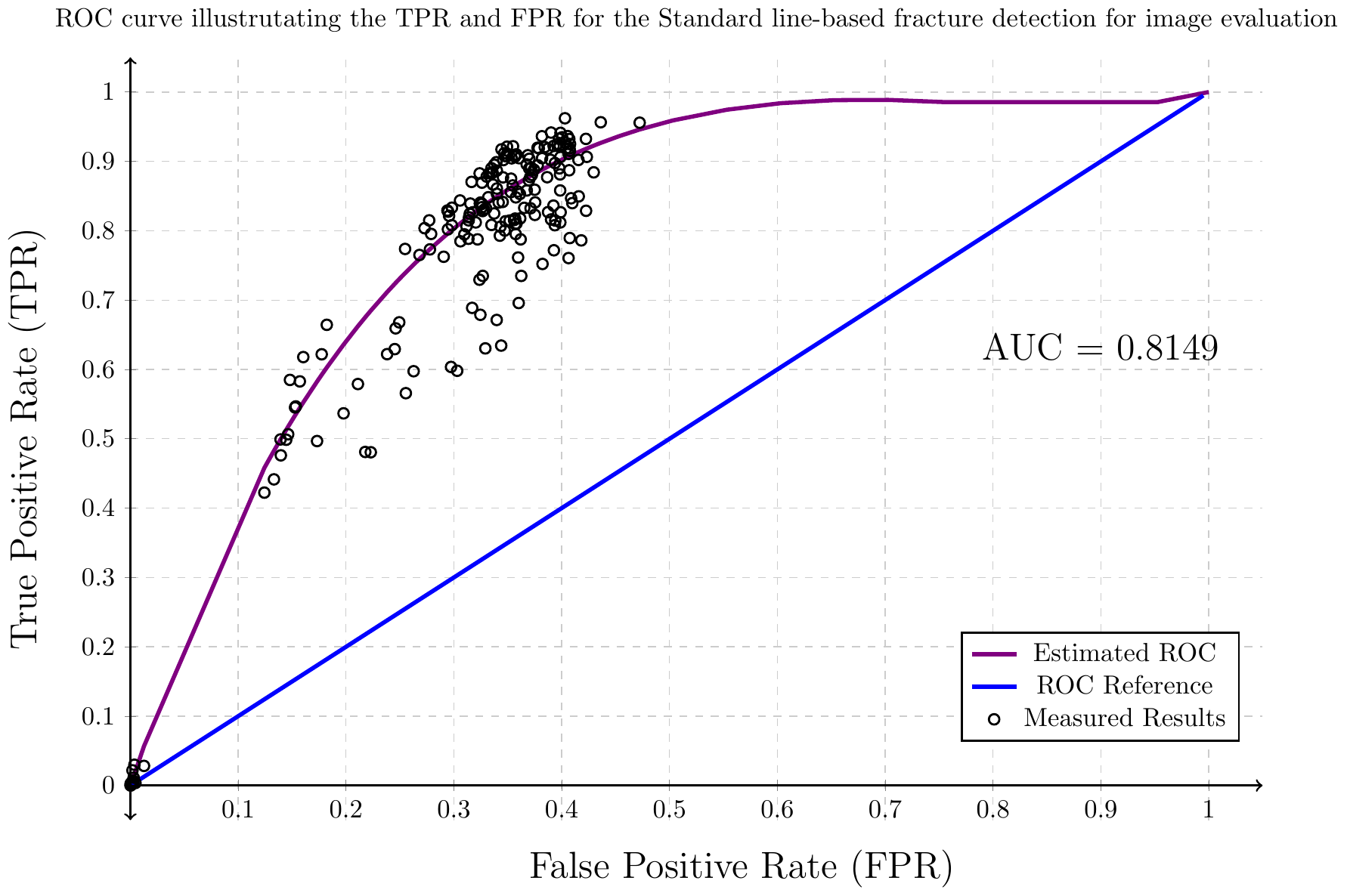}
	\caption{Figure illustrating ROC curve for the Standard line-based fracture detection}
	\label{fig: standard line ROC}
\end{figure}

The ROC curve is constructed using the sensitivity and specificity results from the binary classification system. There are four variables that are considered for the sensitivity and specificity, namely, true positive, false negative, false positive and true negative. Both false negative and false positive are miss-classifications of the lines and have detrimental consequences within the medical field. However, a false negative is far more severe than a false positive, as it implies that a fracture has been missed and can potentially be untreated leading to dire consequences. The calculation for both sensitivity and specificity are expressed in \ref{eq: sensitivity} and \ref{eq: specificity}, respectively. The receiver operating characteristic (ROC) curve is a tool commonly utilised for performance evaluation for binary classification systems \cite{narkhede_understanding_2018}. The ROC curve for the image evaluation is illustrated in Figure \ref{fig: standard line ROC}. The ROC curve is determined by employing curve-fitting on the measured results. The area under the curve (AUC) for the presented ROC graph is 0.8149. An ideal AUC has a value of 1. Therefore, the system has a favourable true positive sensitivity detection. 

\begin{equation}
\label{eq: sensitivity}
sensitvity = \frac{\text{TP}}{\text{TP} + \text{FN}}
\end{equation} 

\begin{equation}
\label{eq: specificity}
specificity = \frac{\text{FP}}{\text{FP} + \text{TN}}
\end{equation}

\begin{figure}[ht!]
	\begin{center}
		\includegraphics[scale=0.8]{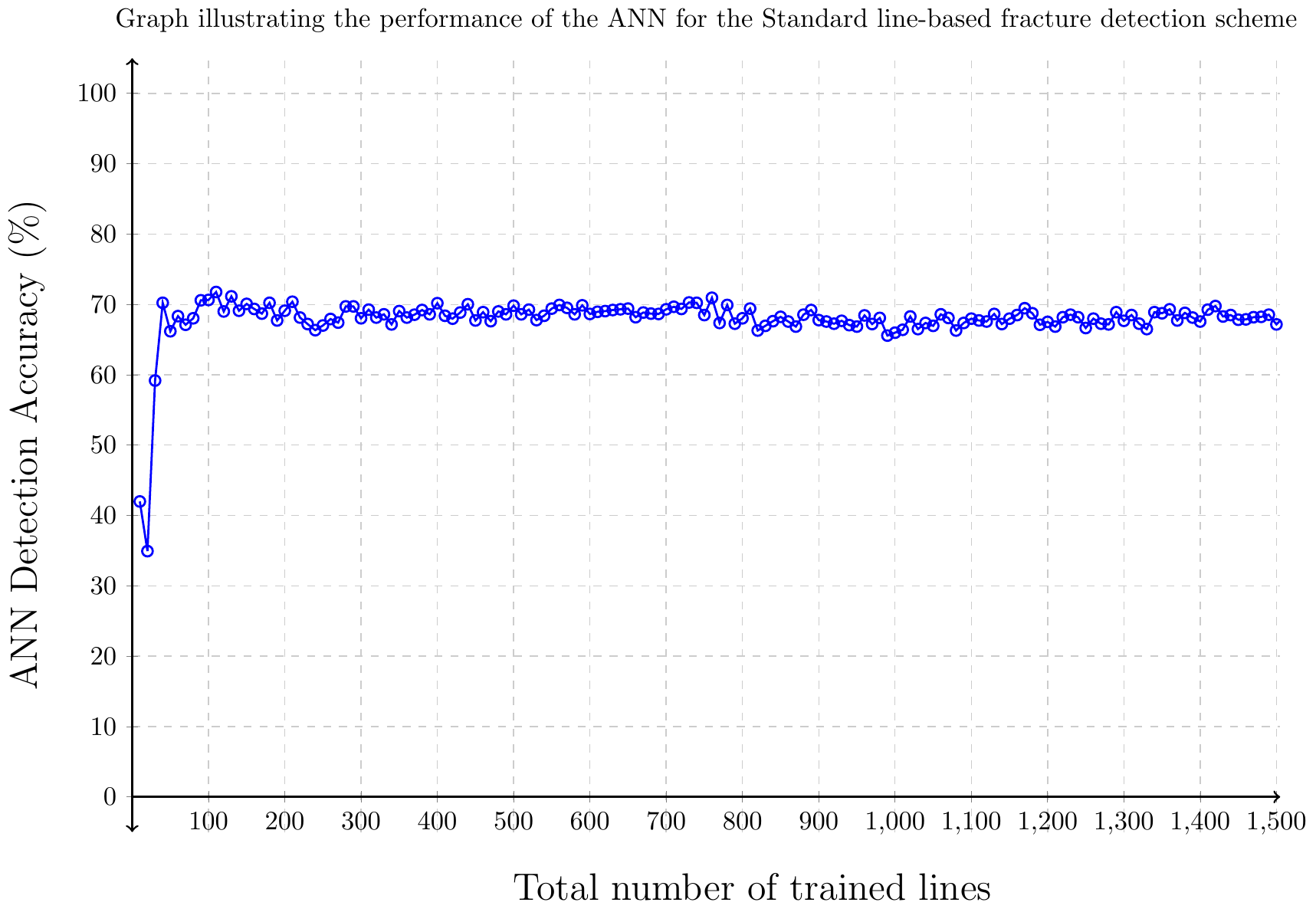}
	\end{center}
	\caption{Graph illustrating the performance of the ANN whilst training it with an equal number of fractured and non-fractured lines for each case for the Standard line-based fracture detection scheme}
	\label{fig: lines accuracy results}
\end{figure}

For the ANN evaluation, each case is made up of a number of lines that is used to train the ANN to evaluate its performance. The lines are grouped into two categories, fractures and non-fractures. The evaluation is set-up such that there is an equal number of fractured and non-fractured lines for each case. Thus 50\% of lines are fractured and 50\% of lines are non-fractured. Each case consists of a number both fractured and non-fractured lines in groups of 5's. The total number of lines used for training is 1,500 lines. Consequently there are a total of 150 cases. The performance of the ANN is evaluated in the same manner as the system evaluation, whereby a fixed number of lines is used for testing the accuracy of the ANN. A total of 13,178 lines are used to test each case, whereby 5,162 are labelled as fractured lines and 8,016 are non-fractured lines. The result of the accuracy for each case is illustrated in Figure \ref{fig: lines accuracy results}.

Table \ref{Table: Line Neural Network Accuracy} shows that despite the number of images trained, it does not affect the accuracy of the ANN as the accuracy of the system ranges between 65.37\% to 75.44\%. This indicates that a single image provides the ANN with enough line training data to achieve an average of 71.57\% accuracy. Figure \ref{fig: lines accuracy results} illustrates that accuracy of the ANN is between 67\% to 72\% after the exposure of 40 lines. Therefore, the minimum number of lines needed to achieve an accuracy between 67\% to 72\% are 20 fractured and 20 non-fractured lines. An increase in the number of lines exposed to the ANN, does not result in any drastic improvements regarding the detection accuracy.

\newpage
\section{Adaptive Differential Parameter Optimized Line-Based Fracture Detection} \label{Section: Optimized Line-Based Fracture Detection}
For the Standard line-based fracture detection scheme, the parameters of the Probabilistic Hough Transform is not optimized for line detection. The purpose of the Adaptive Differential Parameter Optimized (ADPO) line-based fracture detection scheme is to optimize the parameters, such that the generated lines can accurately represent the image edge objects found in the X-ray image. This includes generating granule lines that details the fractures in the fractured region. There are three major parameters that are optimized: threshold, minimum line length and maximum line gap parameters. The ADPO line-based fracture detection scheme follows the same procedure as the Standard scheme. The difference is that the ADPO, introduces a parameter optimizing process and a filtering technique to separate surrounding flesh lines from the leg-bone lines in the leg region. The architecture and training of the ANN remains the same as the Standard scheme. Additionally, the evaluation for both system and ANN remains the same.

\subsection{Adaptive Differential Parameter Optimization} \label{Section: Adaptive Differential Paramenter Optimization}
\begin{enumerate}
	\item The threshold parameter controls the number of point intersections, such that $(r_\theta, \theta)$ coordinates are considered as lines. An increased threshold value defines fewer lines with less minor details about the image edge objects, whereas a decreased threshold parameter value captures all details about the image. This is illustrated in Figure \ref{fig: threshold difference}. Therefore, the threshold parameter value is set to 1. Although the detected lines in Figure \ref{fig: Threshold 1} are scattered compared to the lines in Figure \ref{fig: Threshold 50}, the lines are reconstructed at a later stage when all other parameters are optimized.
	
	\begin{figure}[ht!]
		\centering
		\subfigure[Threshold value 1]{\label{fig: Threshold 1}\includegraphics[width=0.49\textwidth]{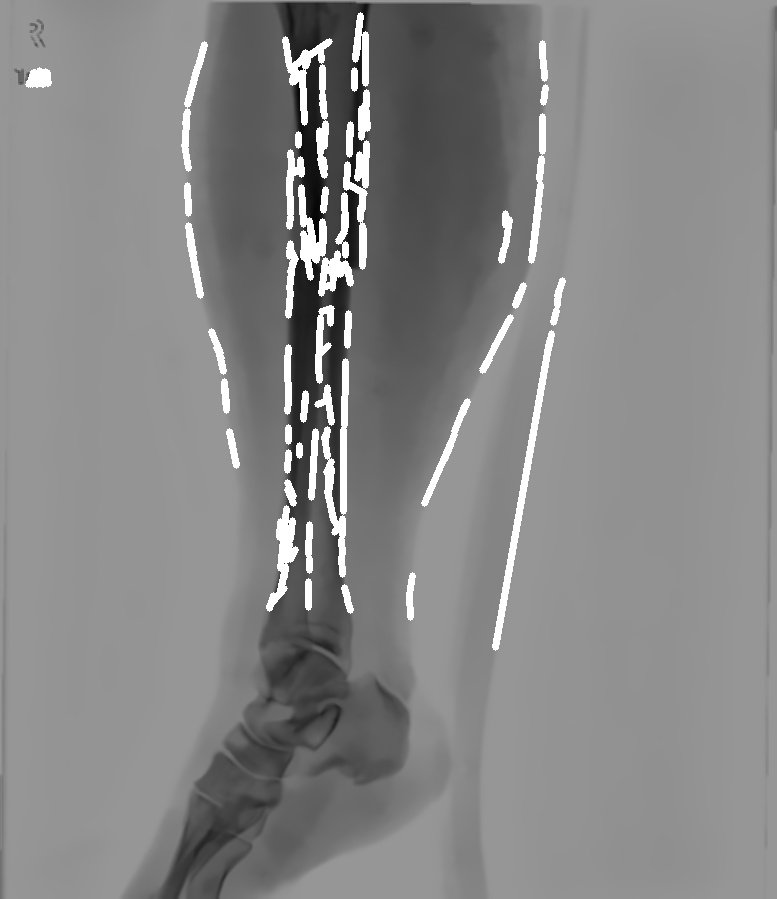}}
		\subfigure[Threshold value 50]{\label{fig: Threshold 50}\includegraphics[width=0.49\textwidth]{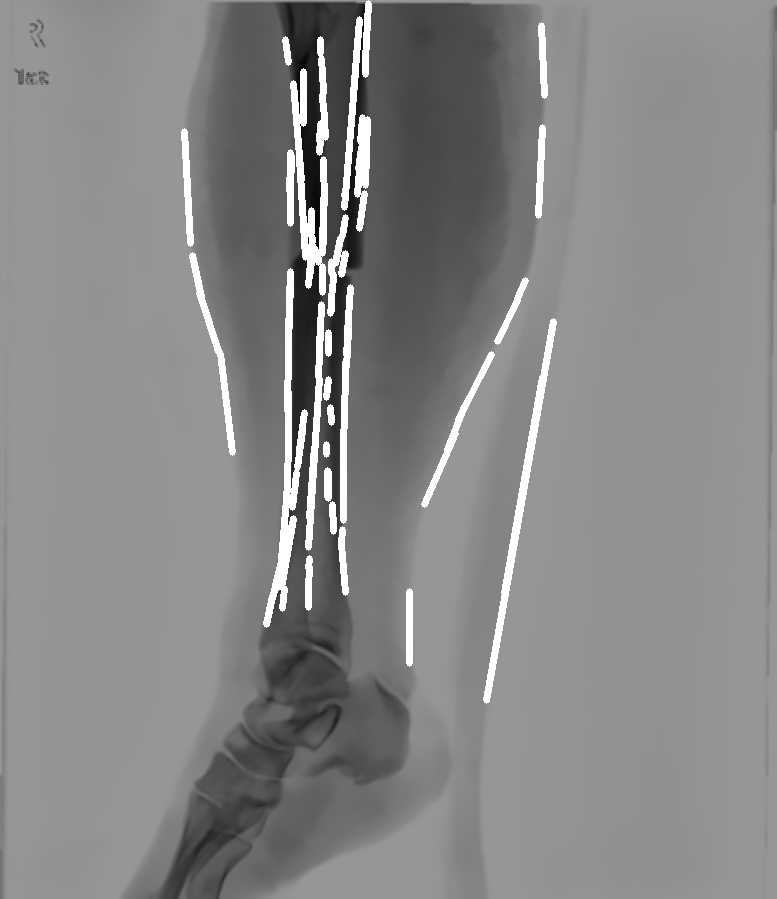}}
		\caption{Images illustrating the generated line difference between a minimal and increased threshold parameter value} 
		\label{fig: threshold difference}
	\end{figure}
	
	\item The minimum line length, $l_{min}$ parameter controls the accepted length of the detected line. An increased $l_{min}$ eliminates lines that are considered as noise, however it disregards detailed lines that are found in the fractured region. This is depicted in Figure \ref{fig: minimum line length difference}, where Figure \ref{fig: minimum line length 2} shows more detailed lines, particularly within the fractured region compared to Figure \ref{fig: minimum line length 25}, whereas in Figure \ref{fig: minimum line length 25}, the outline of the fracture is barely visible from the generated lines. Although a minimal value assigned to $l_{min}$ may be ideal, but the generated lines contains redundant information, as a number of short segmented lines can be represented by a single line instead. Additionally, the shorter lines distorts the most frequently occurring gradient within the leg region and therefore effects the \textit{gradient deviation} feature.
	
	\begin{figure}[ht!]
		\centering
		\subfigure[$l_{min} = 2$]{\label{fig: minimum line length 2}\includegraphics[width=0.49\textwidth]{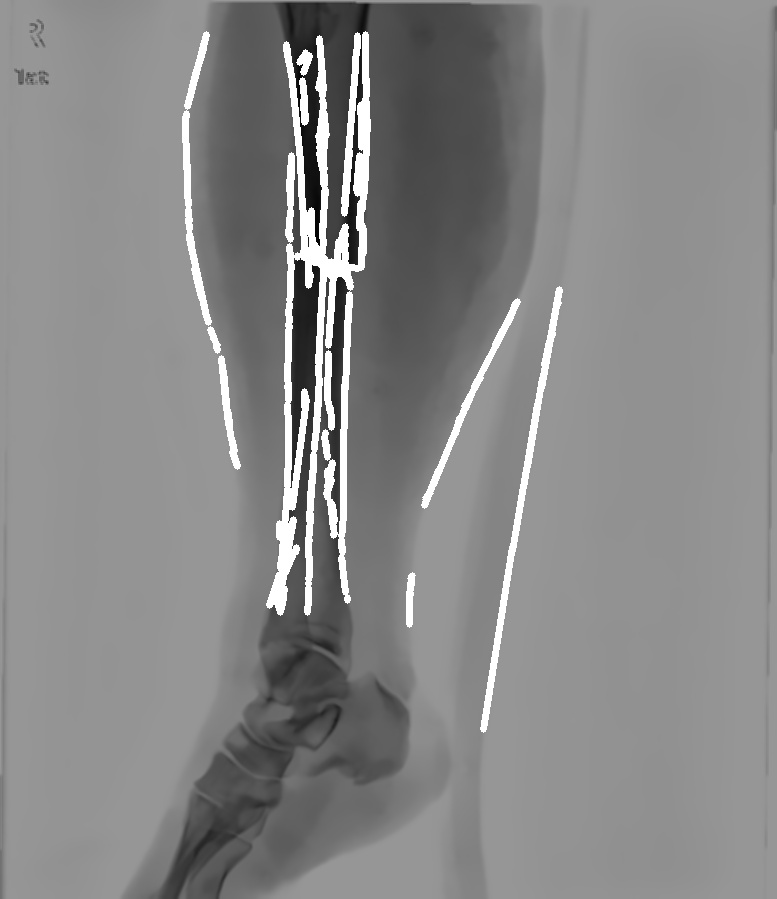}}
		\subfigure[$l_{min} = 25$]{\label{fig: minimum line length 25}\includegraphics[width=0.49\textwidth]{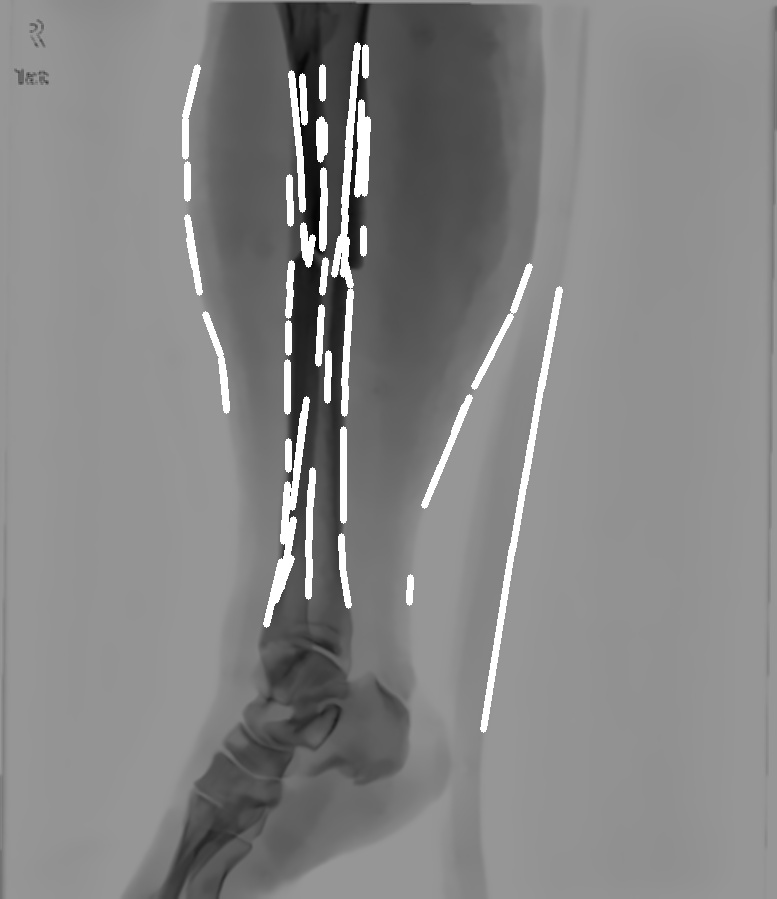}}
		\caption{Images illustrating the generated line difference between a minimal and increased value assigned to the minimum line length parameter} 
		\label{fig: minimum line length difference}
	\end{figure}
	
	The optimized minimum line length, $l'_{min}$ is determined by obtaining the gradients of each line, $L$ detected in the image for a particular $l_{min}$, where $1 \leq l_{min} \leq 25$. Thus there are a total of 25 images generated for each minimum line length ranging from 1 to 25. The gradients of each line within each image is used to calculate the average gradient, $\bar{\theta}_{l_{min}}$ to represent the general direction of the lines in the image. The determination of $\bar{\theta}_{l_{min}}$ is expressed in \eqref{eq: average gradient}. The average gradients are evaluated to determine the maximum different between adjacent average gradients. This evaluation is performed by finding the difference between each adjacent average gradient, which is expressed in \eqref{eq: difference average gradient}. The $l_{min}$ associated to the maximum difference between the average gradients, $\Delta \bar{\theta}_{max}$ is the optimized minimum line length. This is because the gradients of the fractured lines are in a more horizontal position between $0^{\circ}$ and $45^{\circ}$, compared to non-fractured lines which are generally in placed in a vertical position. Therefore, the fractured lines have a lower gradient value than non-fractured lines and the largest average difference is associated with $l'_{min}$. The calculation is expressed in \eqref{eq: average gradient difference}.
	
	\begin{equation}
	\label{eq: average gradient}
	\bar{\theta}_{l_{min}} = \frac{\sum_{k=1}^{m} \theta_{L_k}}{m}
	\end{equation}
	where, $m$ is the total number of detected lines in the image and $L_k$ is the $k$-th line.
	
	\begin{equation}
	\label{eq: difference average gradient}
	\Delta \bar{\theta}_{l_{min}}= \bar{\theta}_{l_{min}} - \bar{\theta}_{l_{min}-1}, \\
	\end{equation}
	
	\begin{equation}
	\label{eq: average gradient difference}
	l'_{min} = argmax(\Delta \bar{\theta}_{l_{min}})
	\end{equation}
	
	However the lines, $\boldsymbol{L}_{l'_{min}}$ generated from the optimized minimum line length does not contain all the detailed lines in the fractured region. Thus, lines generated by prior minimum line lengths from the optimized minimum line length are borrowed to fill in the missing detailed fractured lines. A conditional statement is employed such that the borrowed minimum line length will never be less than the value ``1". The prior minimum line lengths are specifically, $l'_{min} - 1$ and $l'_{min} - 2$. Any further prior lines that are considered are redundant as it shares the same lines with $l'_{min}$, $l'_{min}-1$, and $l'_{min}-2$. The considered lines, $\boldsymbol{L}$ are is described in \eqref{eq: borrowed lines}.
	
	\begin{equation}
	\label{eq: borrowed lines}
	\boldsymbol{L} = \begin{bmatrix}
	\boldsymbol{L}_{l'_{min}} & \boldsymbol{L}_{l'_{min}-1} & \boldsymbol{L}_{l'_{min}-2}
	\end{bmatrix}
	\end{equation}
	where $\boldsymbol{L}$ is a vector of lines, $\text{L}(x_1, y_1, x_2, y_2)$.
	
	There are repeated lines from the three selected sets of line vectors: $\boldsymbol{L}_{l'_{min}}$, $\boldsymbol{L}_{l'_{min}}$, and $\boldsymbol{L}_{l'_{min}}$. Only the unique lines are exacted and considered. Thus, the final lines considered, $\boldsymbol{L}_{f}$ are expressed in \eqref{eq: optimized lines}
	
	\begin{equation}
	\label{eq: optimized lines}
	\boldsymbol{L}_{f}(\boldsymbol{L}) = unique\{L_i \}_{i \in \{1, 2, 3, ..., n \}}
	\end{equation}
	
	\item The maximum line gap, $L^g_{max}$ parameter controls the gap between the ending points of each line segment. If the distance between the line segments' ending points do not meet the allowed maximum line gap, the segments are combined to create a single line. A decreased maximum line gap generates lines that does not describe the image edge objects in the fractured region, despite the optimized minimum line length value. The results are shown in Figure \ref{fig: maximum line gap 10}. An increased maximum line gap generates lines that misrepresents the image edge object, as it over extends the lines. The result of an increased maximum line gap value is shown in Figure \ref{fig: maximum line gap 20}. The selected value for the maximum line gap is 13. The value is chosen based on the evaluation of the number of lines generated for each maximum line gap value ranging between 10 and 20. The maximum line gap value of 13 is chosen as it generates a sufficient number of lines that details the crucial information for 50 images used for testing.  
	\begin{figure}[ht!]
		\centering
		\subfigure[$L^{g}_{max} = 10$]{\label{fig: maximum line gap 10}\includegraphics[width=0.49\textwidth]{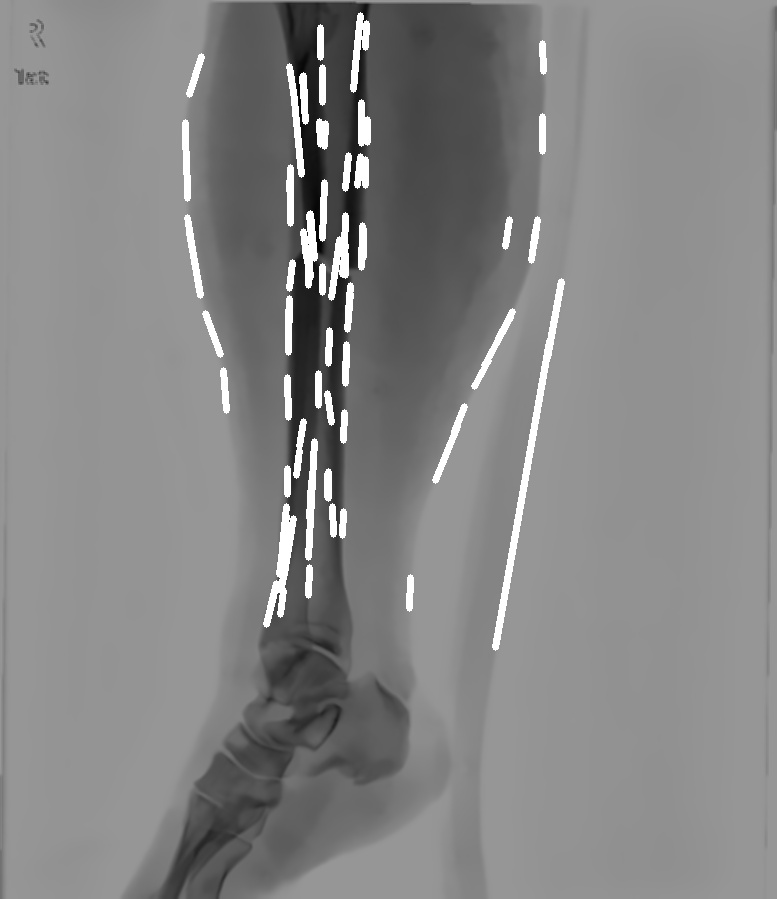}}
		\subfigure[$L^{g}_{max} = 25$]{\label{fig: maximum line gap 20}\includegraphics[width=0.49\textwidth]{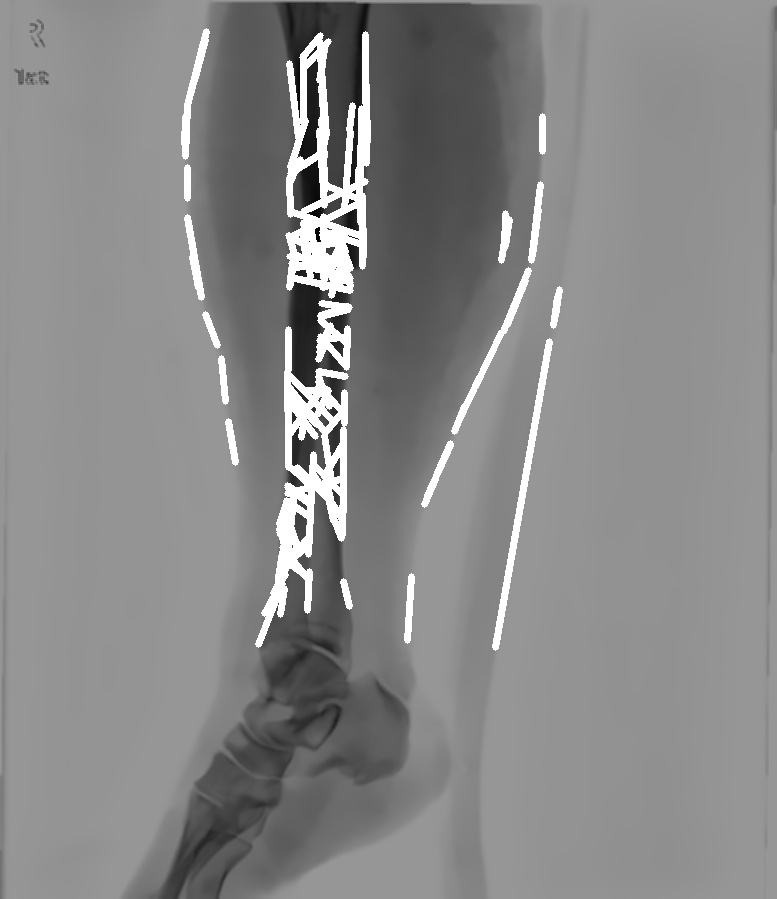}}
		\caption{Images illustrating the generate line difference between a minimal and increased value assigned to the maximum line gap parameter} 
		\label{fig: maximum line gap difference}
	\end{figure}
\end{enumerate}

\subsection{Lines of Interest in the Leg Region} \label{Section: Lines of Interest in the Leg Region}
\begin{figure}[ht!]
	\centering
	\subfigure[All lines detected in leg region]{\label{fig: all lines}\includegraphics[width=0.49\textwidth]{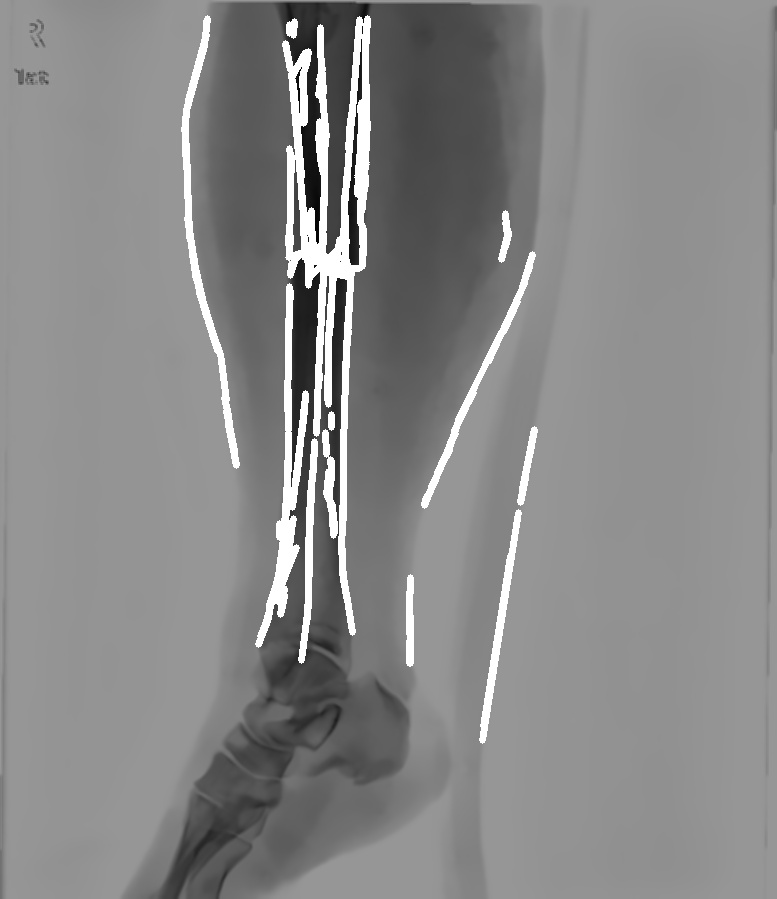}}
	\subfigure[Removal of all other lines except lines in the bone region]{\label{fig: bone lines}\includegraphics[width=0.49\textwidth]{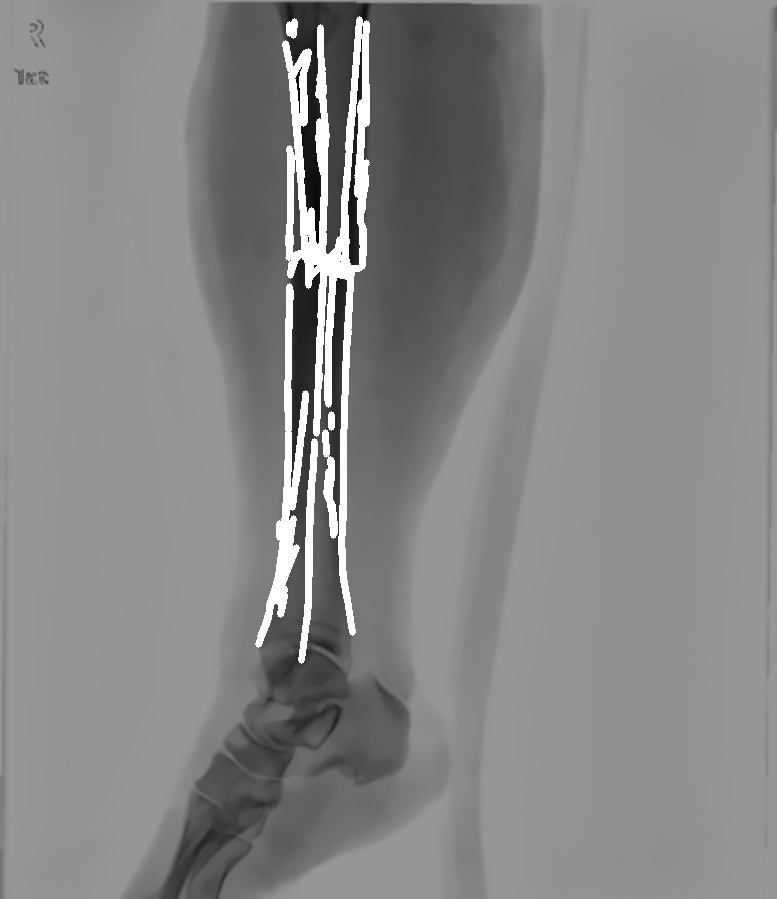}}
	\caption{Images illustrating the before and after of applying the sliding window to isolate the surrounding flesh lines from the leg-bone lines in the leg region} 
	\label{fig: leg region of interest}
\end{figure}

Through visual analysis of lines generated from the X-ray images, the detected lines includes both the lines of the bones and of the surrounding flesh (outer leg) as illustrated in Figure \ref{fig: all lines}. The fractured lines are only situated in the leg-bone region. Therefore, the focus of the fracture detection only concerns the lines within the leg-bone region. Consequently, the surrounding flesh lines are eliminated to reduce the number of lines used for training the ANN. The elimination of the flesh lines is performed through the process of analysing the x-values of each line point. The x-values are chosen for the analysis for two reasons. The first is to isolate the lines in the leg-bone region from the surrounding flesh lines by creating vertical slices of the image. The second reason is due to the pattern discovered with the x-values regarding its difference in density for lines in the flesh and leg-bone region. The leg-bone has a higher x-value density compared to the flesh lines, as there are more detected lines in the leg-bone region. This pattern holds true for all X-ray images used for this research.

In order to focus on the lines in leg-bone region, the lower and upper bounds are determined using the extracted x-values from the lines. Therefore, $\text{x}_1$ and $\text{x}_2$ are extracted from each line, $L(\text{x}_1, \text{y}_1, \text{x}_2, \text{y}_2)$. A frequency vector, $\boldsymbol{f}_x$ for all unique x-values is created. The frequency vector holds the total number of occurrences for each unique x-value in the form of $( x, f )$, where $x$ is the unique x-value and $f$ is the number of occurrences for the particular x-value. The x-values of $\boldsymbol{f}_x$ are analysed by employing a sliding window, $w^{*}$ that traverses through each x-value of the image width, $w$ to find matching values in $\boldsymbol{f}_x$. The window size, $l_{w^{*}}$ of the sliding window is 5\% of the image width, $w$. The 5\% value is selected based on the number of detailed points obtained from the windowing procedure, which is through trial-and-error. An increase in the window size gives leeway for the lower bound value, however the upper bound value is shortened and does not achieve the desired x-value. Thus, by increasing the window size, the ROI is shifted.

\begin{figure}[ht!]
	\centering
	\includegraphics[scale=0.9]{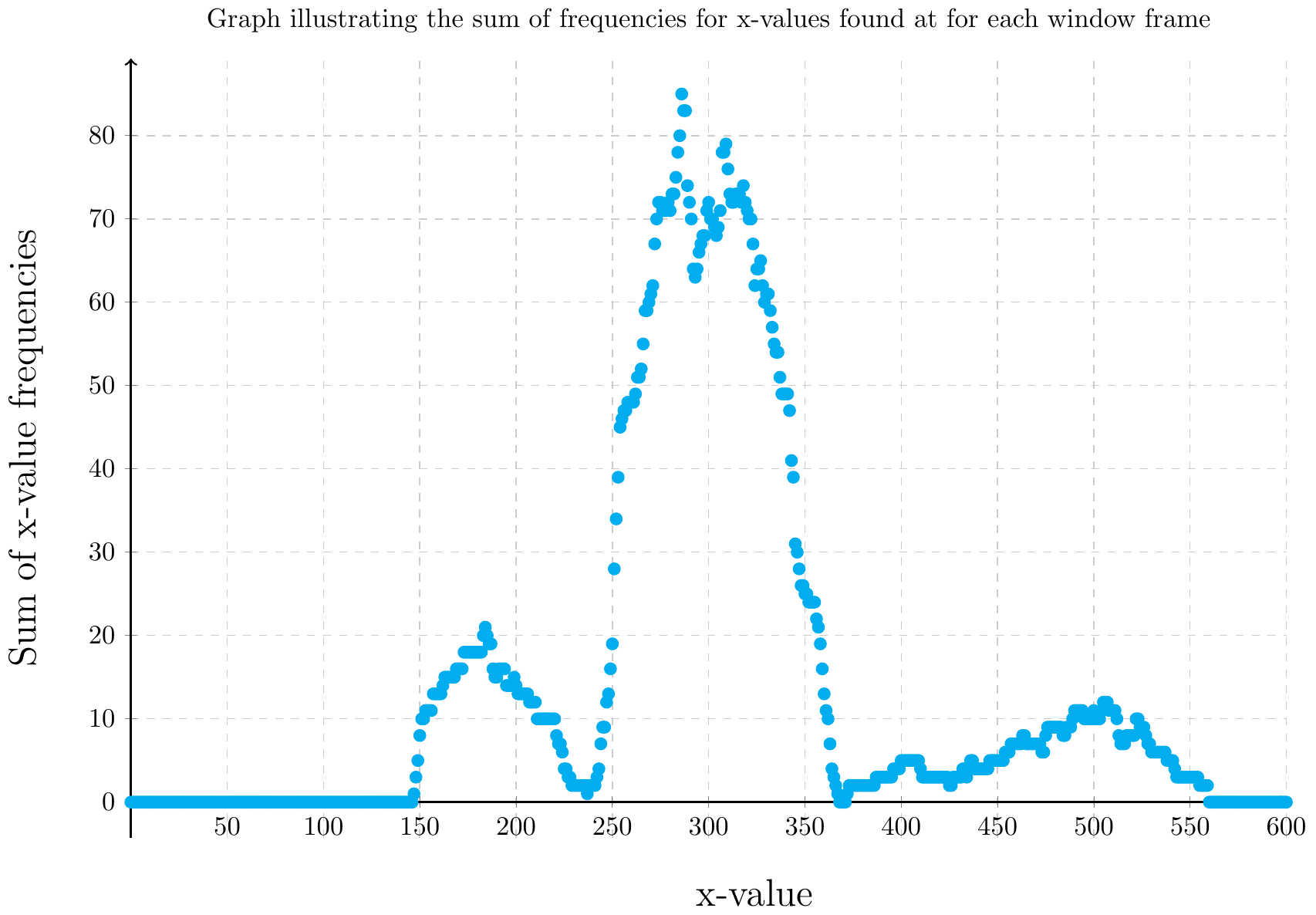}
	\caption{Graph showing the results of the windowing technique to Figure \ref{fig: all lines}}
	\label{fig: windowPointDensity}
\end{figure}

The sliding window is bounded by two variables, the starting window value, $w_s$ and the ending window value, $w_e$. The two variables vary in value as the sliding window traverses through the width of the image, whilst maintaining a fixed length defined by $l_{w^{*}}$. If a match between the $i$-th x-value within the sliding window ($w_s \leq i \leq w_e$) with the $j$-th x-value in $\boldsymbol{f}_x$, then frequency, $f_j$ is summed to obtain the window total frequency, $f^{(i)_{\text{tot}}}$ for the $i$-th width x-value. The purpose for obtaining $f^{(i)}_{\text{tot}}$ for each $i$-th x-value is to find the region with the highest x-value density to determine the lower and upper bounds of the leg-bone region. The described algorithm is presented in Algorithm \ref{alg: density algorithm} and the result of the density algorithm for Figure \ref{fig: all lines} is illustrated in Figure \ref{fig: windowPointDensity}.

\begin{algorithm}[ht!]
	\caption{Obtaining x-Value-Density of x-values Extracted from Lines Algorithm}
	\label{alg: density algorithm}
	\SetAlgoLined 
	\KwData{x-values, $\text{x}_1$ and $\text{x}_2$ from each detected line}
	\KwResult{csv file with summed frequency values at each $i$-th value within the X-ray image}
	Create csv file\;
	Obtain frequency vector $\boldsymbol{f}_x$ for each unique x-value through frequency analysis\;
	Initialise sliding window size, $l_{w^{*}}$ to $5\% $ of image width, $w$\;
	\For{$i = 0$, $i < w$, $i++$}
	{
		Window start, $w_{s} = i$\;
		Window end, $w_{e} = i + l_{w^{*}}$\;
		Window frequency, $f^{(i)}_{\text{tot}} = 0$\;
		
		\For{$j = 0$, $j < l_{\boldsymbol{f}_x}$, $j++$}
		{
			\If{$w_{s} \leq f^{(j)}_x < w_{e}$}
			{$f^{(i)}_{\text{tot}} = f^{(i)}_{\text{tot}} + f_j$\;} 
		}
		record the total frequency within the current window, $f^{(i)}_{\text{tot}}$ along with the associated $i$-th value in the csv file\;
	}
	
	\Return csv file \;
\end{algorithm}

In Figure \ref{fig: windowPointDensity}, the region with the highest peak is indicative of the highest x-value density within the image. The lower and upper bounds of the region is determined by observing the turning points of the graph in Figure \ref{fig: windowPointDensity}. The result of applying the lower and upper bounds is presented in Figure \ref{fig: bone lines}.

\subsection{Optimized Line-Based Detection Results} \label{Section: Optimized Line-Based Detection Results}
The results of the ADPO line-based fracture detection scheme are obtained from the same experimental set-up described in Section \ref{Section: ANN Experiement set-up}. However, the difference is the data used for both training and testing the ANN. The ANN is only trained with lines that are found within the leg-bone region. The elimination of the surrounding flesh lines reduces the training complexity and of the ANN. Therefore, the evaluation of the ANN is focused on classifying fractured lines from non-fractured lines found only in the leg-bone region, whilst all other lines are evaluated by other components in the system. This means that the lines found within the knee, foot and flesh regions are automatically classified as non-fractured lines. A total of 16,515 lines extracted from 23 images are employed to evaluate the ADPO scheme. Of the 16,515 lines, 8,035 are fractured lines and 8,480 are non-fractured lines.

\begin{figure}[ht!]
	\centering
	\includegraphics[scale=0.8]{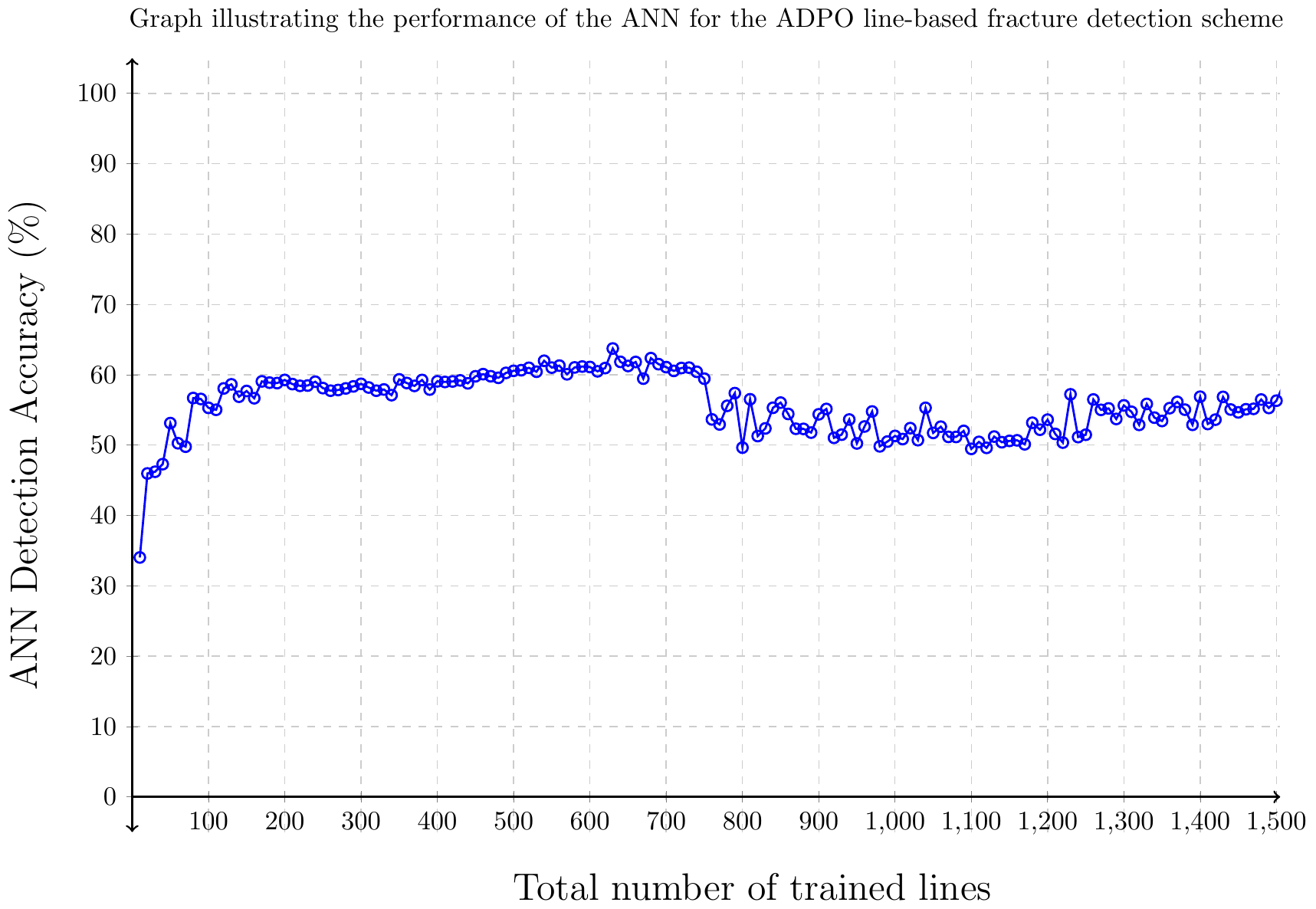}
	\caption{Graph showing the system's minimum, average and maximum accuracies of the ADPO line-based fracture detection scheme}
	\label{fig: optimized line evaluation accuracy}
\end{figure}

The image evaluation accuracy results for the ADPO line-based fracture detection scheme is presented in Table \ref{Table: Optimized Line Neural Network Accuracy}. The average number of lines per image used for the ADPO scheme is 849 lines per image. The accuracy of the system ranges from 63.61\% to 80.88\%. Thus yielding an average accuracy of 72.89\%, which is a slight improvement from the average accuracy obtained from the Standard scheme, which is 71.57\%. Figure \ref{fig: optimized line evaluation accuracy} illustrates the performance of the ANN. A total of 11,195 lines are used to test each case for the ANN, whilst 8,035 lines are fractured and 3,160 are non-fractured. The results shows that the accuracy of the ANN requires a minimum of 300 lines to obtain a 65\% accuracy. The accuracy achieved for the ANN evaluation is less than the accuracy of the Standard scheme, which obtained a maximum accuracy of 72\%. However, the ANN of the ADPO scheme focuses only on the lines in leg-bone region, whereas the ANN of the Standard scheme is exposed to all lines detected in the X-ray image.

\begin{table}[ht!]
	\centering
	\caption{The results for the system's minimum, average and maximum accuracies for 20 cases over 10 simulations for the ADPO line-based fracture detection scheme}
	\label{Table: Optimized Line Neural Network Accuracy}
	\begin{tabular}{c c c c}
		\hline
		\textbf{No. Trained Images} & \textbf{Min Accuracy (\%)} & \textbf{Average Accuracy (\%)} & \textbf{Max Accuracy (\%)} \\
		\hline 
		1 & 65.129 & 73.8656 & 80.866 \\
		2 & 63.294 & 74.7085 & 82.894 \\
		3 & 62.246 & 72.0613 & 81.732 \\
		4 & 63.53 & 72.3312 & 80.751 \\
		5 & 61.629 & 74.2942 & 81.483 \\
		6 & 61.32 & 70.5336 & 80.715 \\
		7 & 65.389 & 74.9762 & 81.163 \\
		8 & 63.76 & 70.1397 & 81.998 \\
		9 & 65.722 & 75.3762 & 81.417 \\
		10 & 62.168 & 70.9889 & 80.648 \\
		11 & 62.313 & 71.7632 & 81.217 \\
		12 & 61.508 & 72.1551 & 80.587 \\
		13 & 65.752 & 74.2622 & 82.955 \\
		14 & 64.087 & 71.7565	& 78.401 \\
		15 & 59.891	& 72.1895 &	80.793 \\
		16 & 65.934	& 71.1875 & 77.312 \\
		17 & 60.539	& 73.1001 & 79.752 \\
		18 & 68.314 & 75.1457 & 81.695 \\
		19 & 64.693 & 72.3942 & 80.339 \\
		20 & 65.05 & 74.5831 & 80.793 \\
		\hline
	\end{tabular}
\end{table}

\begin{figure}[ht!]
	\centering
	\includegraphics[scale=0.8]{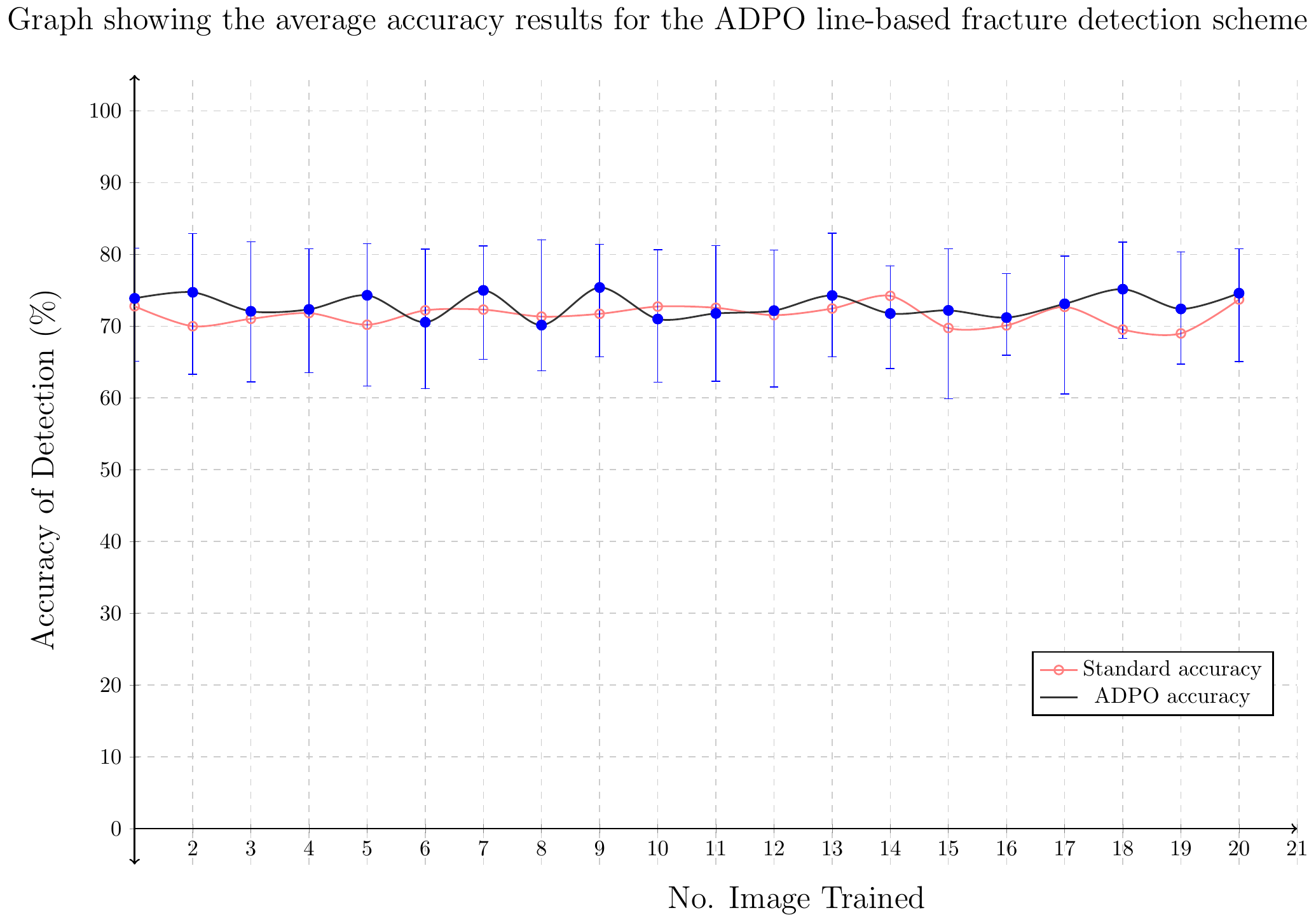}
	\caption{Graph illustrating the average accuracy for 20 cases over 10 simulations for the improved ADPO line-based fracture detection scheme}
	\label{fig: ADPO line accuracy graph with error bars}
\end{figure}

\begin{figure}[ht!]
	\centering
	\includegraphics[scale=0.75]{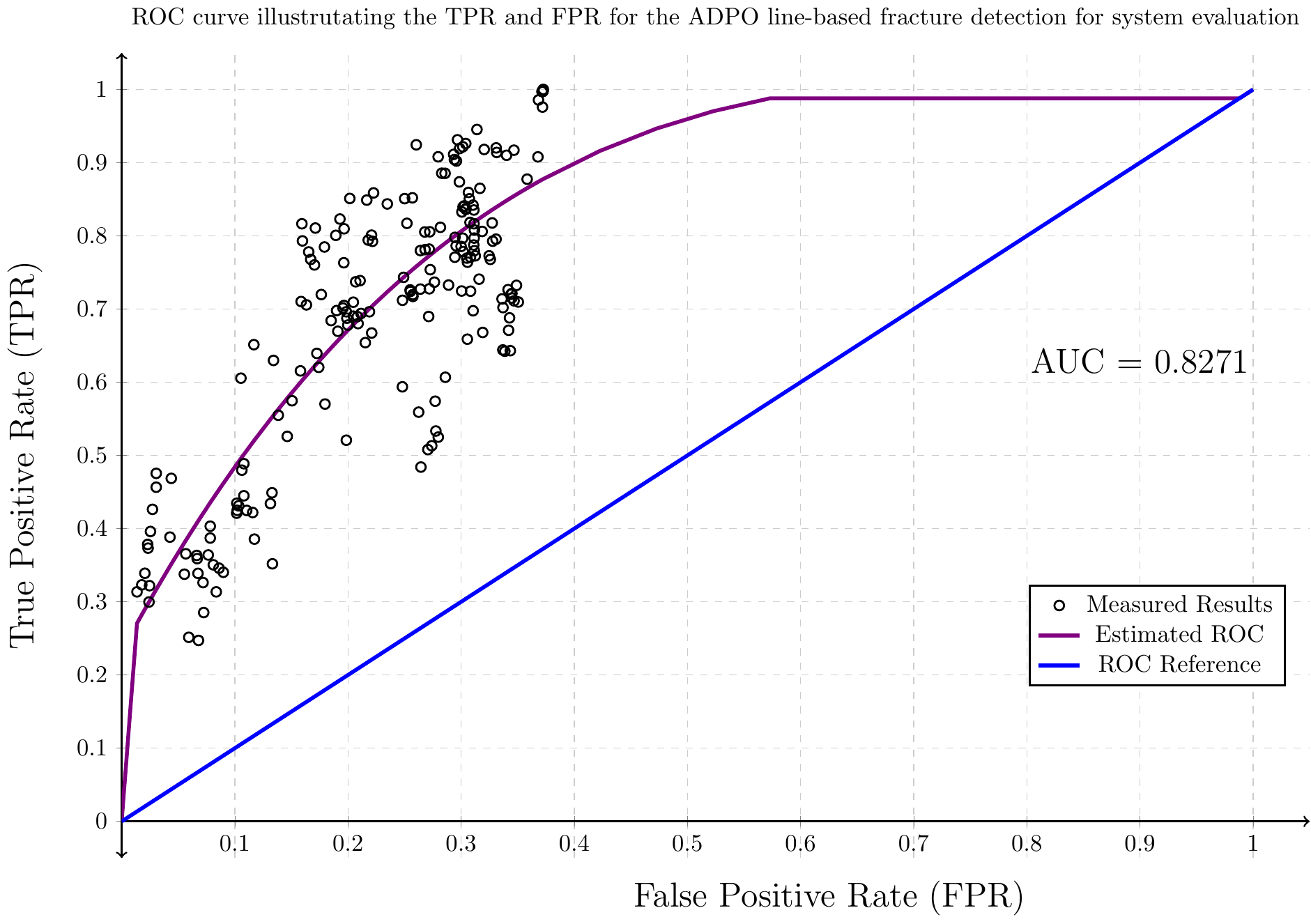}
	\caption{Figure illustrating ROC curve for the ADPO line-based fracture detection}
	\label{fig: ADPO line ROC}
\end{figure}

\newpage
The ROC curve for the system evaluation is presented in Figure \ref{fig: ADPO line ROC}. There are three distinct ROC results illustrated: measured results, estimated ROC, and ROC reference. The measured results are obtained from each case used to evaluate the system. Since the measured results are scattered, a ROC approximation using the measured results is constructed. The estimated ROC curve is employed for the calculation of the AUC, in which the AUC has a value of 0.8271. The ROC reference provides further context of the system's performance relative to a system that randomises binary classification. The AUC value obtained shows that there is  a slight improvement from the Standard scheme, where the AUC value is 0.8149. Therefore, the ADPO scheme is more sensitive in detecting true positives than the Standard scheme.

\newpage
\section{Critical Analysis}
The performance of the ANN is evaluated based on its ability to accurately classify fractured and non-fractured lines. There are four variables that are employed to evaluate the results of both the ANN and the entirety of the system: true positive, true negative, false positive and false negative.

In the medical field, false positives are tolerated, as it is indicative of a cautious system. However, false negatives cannot not be tolerated as it means that a fracture is being detected as normal. It is difficult to avoid false negatives when working with ANN's due to its purpose for forming a generalised operation for all input data. Therefore, the number of false negatives must be as minimal as possible. The accuracy, $a$ for the evaluation of both the ANN and the system is expressed in \eqref{eq: accuracy equation}.

\begin{equation}
\label{eq: accuracy equation}
a = \frac{\text{TP} + \text{TN}}{\text{TP} + \text{TN} + \text{FP} + \text{FN}}
\end{equation}

\begin{figure}[ht!]
	\centering
	\includegraphics[scale=0.8]{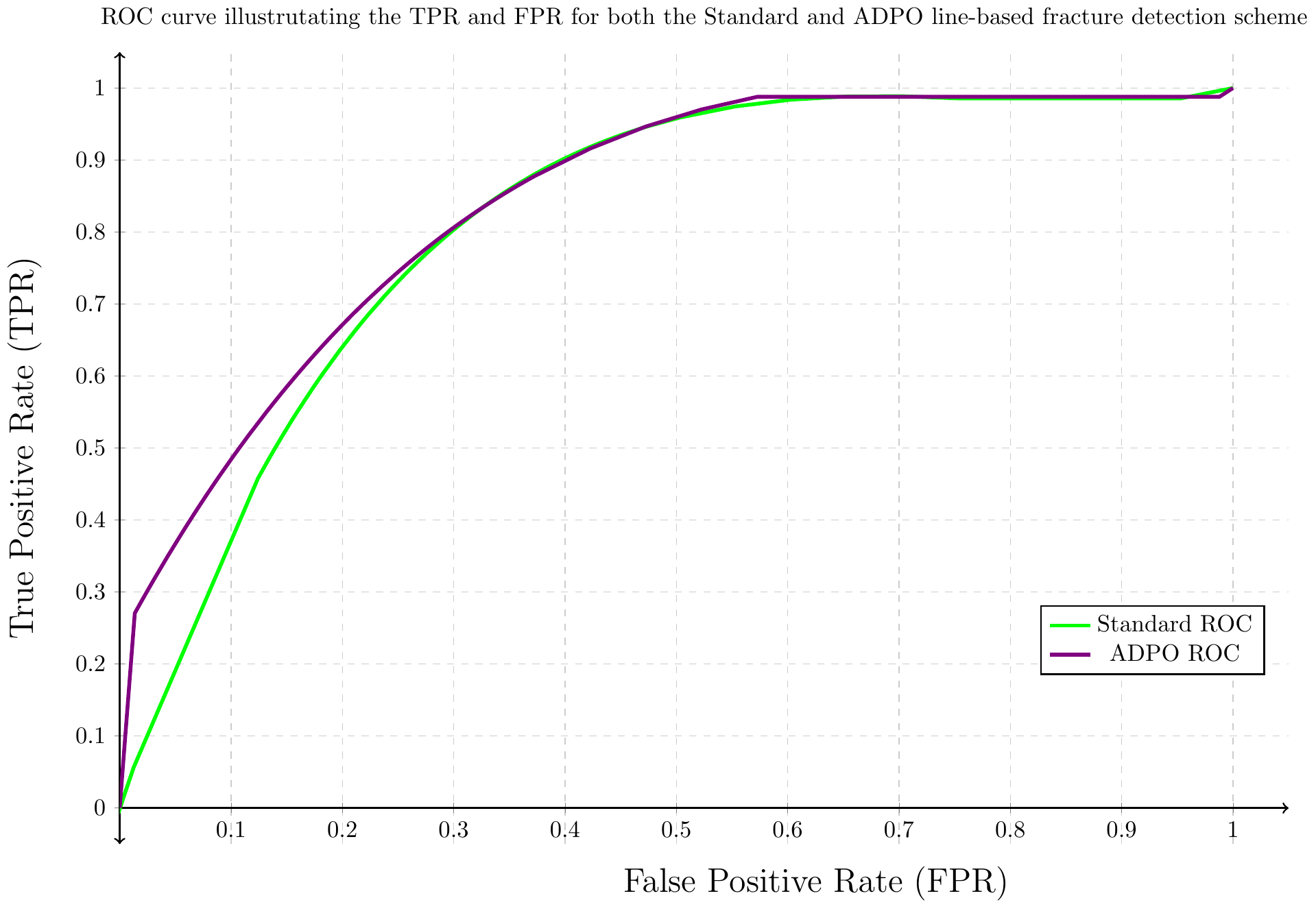}
	\caption{ROC curves for both the Standard and ADPO line-based fracture detection scheme}
	\label{fig: combined ROC}
\end{figure}

Further evaluation of the ROC curves for both Standard and ADPO schemes in Figure \ref{fig: combined ROC}, the ROC curve of the ADPO scheme increases in TPR at a lower FPR value compared to the ROC curve of the Standard scheme. The ROC graph of Figure \ref{fig: standard line ROC} indicates that the scheme reaches maximum sensitivity at 0.4 FPR for the measured results, whilst in Figure \ref{fig: ADPO line ROC} the maximum sensitivity is reached at 0.37 FPR. This indicates that the sensitivity performance of the ADPO scheme is slightly better than the Standard scheme.

Although, the ADPO scheme has a better sensitivity and accuracy performance compared to the Standard scheme, the system's maximum achievable accuracy is 82.9\%, which can be further improved. However, the accuracy of the described schemes are hindered by the labelling of the lines. The approach utilised for the labelling of the fractured and non-fractured lines is an area selection approach, whereby the lines that are within the selected area are labelled as fractures. Lines that are not considered as fractures by human visualisation but are within the selected area are mislabelled. This affects the ANN performance as the weights in the ANN are adjusted due to the mislabelled lines. However, the affect on the weights is minimal as there is only a minority of mislabelled lines. Additionally, the number of features extracted from the lines is limited, because only two points are provided by the Probabilistic Hough Transform to describe the line detected from the image edge objects of the X-ray image. Moreover, the line detection within the X-ray image is only an approximation of the image edge objects in the X-ray image. An improved detection methodology built from the foundations of the line-based fracture detection scheme, whereby contours are utilised rather for feature extraction instead of lines.

\section{Future Improvements}
Future improvements can be implemented to the labelling process of fractured and non-fractured lines, as the current labelling process utilises an area selection approach which mislabels non-fractured lines as fractured lines when the lines are in the area of selection. The improvement to the labelling process includes a deselection process. Thus allowing the user to deselect individual lines from the selected group of fractured lines. This improvement incorporates both area selection and individual deselection. Other improvements is the use of contours over lines, as the detected lines are limited in the number of features that are extracted. Additionally, the lines are an approximation of the edge image objects, whereas contours are a more detailed representation of the edge image objects and holds more points, which result in additional features that can be extracted.

\section{Conclusion}
To conclude, this paper details two line-based fracture detection schemes, Standard and ADPO. Both line-based fracture detection schemes utilises 13 extracted features from the detected lines. The difference is that the ADPO scheme optimises the parameters in the Probabilistic Hough Transform for line detection. The optimisation detects more detailed lines for the fractured region within the X-ray image compared to the Standard scheme. Additionally, the ADPO scheme employs a technique which isolates the bones within the leg region from the surrounding lines of the flesh. Therefore, the ANN is only trained with lines within the leg-bone regions. Thus, shifting the focus of the ANN to detect fractures within the leg-bone area only. The system eliminates all other detected lines, which are automatically classified as non-fractures. This includes lines within the knee, foot and surrounding flesh region. In focusing the ANN for lines only in the leg-bone region, it reduces the training complexity of the ANN. The accuracy of the ADPO scheme is slightly better than the accuracy of the Standard scheme. The average accuracy of the ADPO system is 72.89\%, whilst the average accuracy of the Standard scheme is 71.57\%. 

The ANN evaluation without image context indicates that the Standard scheme performs better than the ADPO scheme, however the ADPO ANN only focuses on the lines found within the leg-bone region, whereas the Standard scheme's ANN is given all detected lines for training. Therefore, the accuracy evaluation will be higher for the Standard scheme as there is a larger number of classified true negatives in the accuracy calculation compared to the ADPO scheme. Further analysis is performed on both schemes to evaluate the ratio of false positive to false negative detection. The ADPO scheme has a higher false negative detection compared to the Standard scheme, but the sensitivity of the ADPO scheme is  better than the sensitivity of the Standard scheme. This is determined by employing ROC curves and the calculation of the AUC. The AUC of the Standard scheme is 0.8149, whilst the AUC of the ADPO scheme is 0.8271. Therefore, between the two line-based fracture detection schemes, the ADPO scheme is preferred over the Standard scheme. 

The limitation of both schemes lies within the detection of the lines, where the detected lines are an approximation of the image edge objects in the X-ray image. Additionally, the two points provided to describe the lines limits the number of features that are extracted for fracture detection. Due to the limitations of line detection and extracted features an alternative approach is considered. The alternative approach employs the use of contours over the use of lines for feature extraction.

\section*{Acknowledgements}

Acknowledgements are extended to the Department of Radiology at No. 85 Hospital, Shanghai, China for the re-usage of the provided X-ray images.

\bibliographystyle{IEEEtran}
\bibliography{mybibfile}

\end{document}